\documentclass[10pt,twocolumn,letterpaper]{article}

\usepackage{cvpr}
\usepackage{times}
\usepackage{epsfig}
\usepackage{graphicx}
\usepackage{amsmath}
\usepackage{amssymb}
\usepackage{comment}
\usepackage{xcolor}
\usepackage{footnote}
\usepackage{subcaption}
\usepackage{mathtools}
\usepackage[]{algorithm2e}
\usepackage{stfloats}
\usepackage{multirow}

\usepackage[pagebackref=true,breaklinks=true,letterpaper=true,colorlinks,bookmarks=false]{hyperref}

\cvprfinalcopy 

\newcommand{\KEYPOINTS}{97607}
\newcommand{\OBJECTS}{14225}
\newcommand{\MODELS}{3049}
\newcommand{\SCANS}{1506}
\newcommand{\OUTPERFORM}{$21.39\%$}

\def\cvprPaperID{977} 

\ifcvprfinal\fi
\begin{document}

\newcommand{\OURS}{Scan2CAD}
\title{\vspace{-1.5cm}\OURS: Learning CAD Model Alignment in RGB-D Scans\vspace{-0.5cm}}

\newcommand{\footremember}[2]{%
   \footnote{#2}
    \newcounter{#1}
    \setcounter{#1}{\value{footnote}}%
}
\newcommand{\footrecall}[1]{%
    \footnotemark[\value{#1}]%
}

\author{
Armen Avetisyan$^{1}$ \qquad Manuel Dahnert$^{1}$  \qquad Angela Dai$^{1}$ \qquad Manolis Savva$^{2}$ 
\vspace{0.1cm} \\ 
Angel X. Chang$^{2}$  \qquad Matthias Nie{\ss}ner$^{1}$
\vspace{0.2cm} \\ 
$^{1}$Technical University of Munich \qquad $^{2}$Simon Fraser University 
}

\twocolumn[{%
	\renewcommand\twocolumn[1][]{#1}%
	\maketitle
	\begin{center}
 		\vspace{-0.8cm}
		\includegraphics[clip,width=0.85\linewidth]{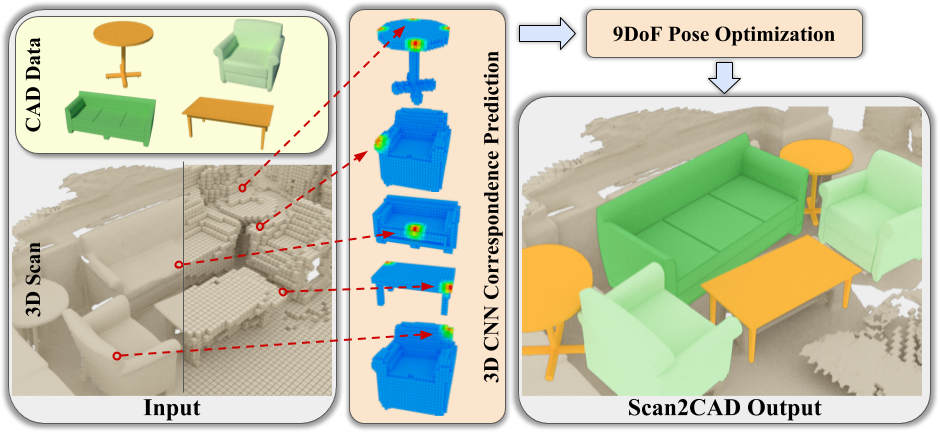}
	 \end{center}
	 	\vspace{-0.3cm}
	    \captionof{figure}{\OURS{} takes as input an RGB-D scan and a set of 3D CAD models (left).
	    We then propose a novel 3D CNN approach to predict heatmap correspondences between the scan and the CAD models (middle).
		From these predictions, we formulate an energy minimization to find optimal 9 DoF object poses for CAD model alignment to the scan (right).}
    	\label{fig:teaser}
	\vspace{0.5cm}
}]

\begin{abstract}
We present \OURS{}\footnote{The Scan2CAD dataset is publicly released along with an automated benchmark script for testing under \url{www.Scan2CAD.org}}, a novel data-driven method that learns to align clean 3D CAD models from a shape database to the noisy and incomplete geometry of an RGB-D scan.
For a 3D reconstruction of an indoor scene, our method takes as input a set of CAD models, and predicts a 9DoF pose that aligns each model to the underlying scan geometry.
To tackle this problem, we create a new scan-to-CAD alignment dataset based on \SCANS{} ScanNet scans with \KEYPOINTS{} annotated keypoint pairs between \OBJECTS{} CAD models from ShapeNet and their counterpart objects in the scans.
Our method selects a set of representative keypoints in a 3D scan for which we find correspondences to the CAD geometry.
To this end, we design a novel 3D CNN architecture to learn a joint embedding between real and synthetic objects, and thus predict a correspondence heatmaps.
Based on these correspondence heatmaps, we formulate a variational energy minimization that aligns a given set of CAD models to the reconstruction.
We evaluate our approach on our newly introduced \OURS{} benchmark where we outperform both handcrafted feature descriptor as well as state-of-the-art CNN based methods by \OUTPERFORM{}.
\end{abstract}

\vspace{-0.3cm}
\section{Introduction}

In recent years, the wide availability of consumer-grade RGB-D sensors, such as the Microsoft Kinect, Intel Real Sense, or Google Tango, has led to significant progress in RGB-D reconstruction.
We now have 3D reconstruction frameworks, often based on volumetric fusion~\cite{curless1996volumetric}, that achieve impressive reconstruction quality~\cite{izadi2011kinectfusion,newcombe2011kinectfusion,niessner2013hashing,whelan2015elasticfusion,kahler2015very} and reliable global pose alignment~\cite{whelan2015elasticfusion,choi2015robust,dai2017bundlefusion}.
At the same time, deep learning methods for 3D object classification and semantic segmentation have emerged as a primary consumer of large-scale annotated reconstruction datasets~\cite{dai2017scannet,Matterport3D}.
These developments suggest great potential in the future of 3D digitization, for instance, in applications for virtual and augmented reality.

Despite these improvements in reconstruction quality, the geometric completeness and fine-scale detail of indoor scene reconstructions remain a fundamental limitation.
In contrast to artist-created computer graphics models, 3D scans are noisy and incomplete, due to sensor noise, motion blur, and scanning patterns.
Learning-based approaches for object and scene completion~\cite{dai2017complete,song2017ssc,dai2018scancomplete} cannot reliably recover sharp edges or planar surfaces, resulting in quality far from artist-modeled 3D content.

One direction to address this problem is to retrieve a set of CAD models from a shape database and align them to an input scan, in contrast to a bottom-up reconstruction of the scene geometry.
If all objects are replaced in this way, we obtain a clean and compact scene representation, precisely serving the requirements for many applications ranging from AR/VR scenarios to architectural design.
Unfortunately, matching CAD models to scan geometry is an extremely challenging problem:
While high-level geometric structures might be similar, the low-level geometric features differ significantly (e.g., surface normal distributions).
This severely limits the applicability of handcrafted geometric features, such as FPFH~\cite{rusu2009fast}, SHOT~\cite{salti2014shot}, point-pair-features~\cite{drost20123d}, or SDF-based feature descriptors~\cite{li2015database}. 
While learning-based approaches like random forests~\cite{nan2012search,shao2012interactive} exist, their model capacity remains relatively low, especially in comparison to more modern methods based on deep learning, which can achieve significantly higher accuracy, but remain at their infancy. 
We believe this is in large part attributed to the lack of appropriate training data.

In this paper, we make the following contributions:
\vspace{-0.2cm}
\begin{itemize}
    \item We introduce the \OURS{} dataset, a large-scale dataset comprising \KEYPOINTS{} pairwise keypoint correspondences and 9DoF alignments between \OBJECTS{} instances of \MODELS{} unique synthetic models, between ShapeNet~\cite{shapenet2015} and reconstructed scans in ScanNet~\cite{dai2017scannet}, as well as oriented bounding boxes for each object.
\vspace{-4pt}
    \item We propose a novel 3D CNN architecture that learns a joint embedding between real and synthetic 3D objects to predict accurate correspondence heatmaps between the two domains.
\vspace{-4pt}
    \item We present a new variational optimization formulation to minimize the distance between scan keypoints and their correspondence heatmaps, thus obtaining robust 9DoF scan-to-CAD alignments. 
\end{itemize}

\section{Related work}
\paragraph{RGB-D Scanning and Reconstruction}
The availability of low-cost RGB-D sensors has led to significant research progress in RGB-D 3D reconstruction.
A very prominent line of research is based on volumetric fusion~\cite{curless1996volumetric}, where depth data is integrated in a volumetric signed distance function.
Many modern real-time reconstruction methods, such as KinectFusion~\cite{izadi2011kinectfusion,newcombe2011kinectfusion}, are based on this surface representation.
In order to make the representation more memory-efficient, octree~\cite{chen2013scalable} or hash-based scene representations have been proposed~\cite{niessner2013hashing,kahler2015very}.
An alternative fusion approach is based on points~\cite{keller2013real}; the reconstruction quality is slightly lower, but it has more flexibility when handling scene dynamics and can be adapted on-the-fly for loop closures~\cite{whelan2015elasticfusion}.
Very recent RGB-D reconstruction frameworks combine efficient scene representations with global pose estimation~\cite{choi2015robust}, and can even perform online updates with global loop closures~\cite{dai2017bundlefusion}.
A closely related direction to ours (and a possible application) is recognition of objects as a part of a SLAM method, and using the retrieved objects as part of a global pose graph optimization~\cite{salas2013slam++,mccormac2018fusion++}.

\paragraph{3D Features for Shape Alignment and Retrieval}

Geometric features have a long-established history in computer vision, such as Spin Images~\cite{johnson1997spin}, Fast Point Feature Histograms (FPFH)~\cite{rusu2009fast}, or Point-Pair Features (PPF)~\cite{drost20123d}.
Based on these descriptors or variations of them, researchers have developed shape retrieval and alignment methods.
For instance, Kim et al.~\cite{kim2012acquiring} learn a shape prior in the form of a deformable part model from input scans to find matches at test time; or AA2h~\cite{kim2013guided} use a similar approach to PPF, where a histogram of normal distributions of sample points is used for retrieval. 
Li et al.~\cite{li2015database} propose a formulation based on a hand-crafted TSDF feature descriptor to align CAD models in real-time to RGB-D scans.
While these retrieval approaches based on hand-crafted geometric features show initial promise, they struggle to generalize matching between the differing data characteristics of clean CAD models and noisy, incomplete real-world data.

An alternative direction is learned geometric feature descriptors.
For example, Nan et al.~\cite{nan2012search} use a random decision forest to classify objects on over-segmented input geometry from high-quality scans.
Shao et al.~\cite{shao2012interactive} introduce a semi-automatic system to resolve segmentation ambiguities, where a user first segments a scene into semantic regions, and then shape retrieval is applied.
3DMatch~\cite{zeng20173dmatch} leverage a Siamese neural network to match keypoints in 3D scans for pose estimation. 
Zhou et al.~\cite{zhou2018unsupervised} is of similar nature, proposing a view consistency loss for 3D keypoint prediction network on RGB-D image data. 
Inspired by such approaches, we develop a 3D CNN-based approach targeting correspondences between the synthetic domain of CAD models and the real domain of RGB-D scan data.

Other approaches retrieve and align CAD models given single RGB~\cite{lim2013parsing,izadinia2017im2cad,sun2018pix3d,huang2018holistic} or RGB-D~\cite{gupta2015aligning,zou2018complete} images. 
These methods are related, but our focus is on geometric alignment independent of RGB information, rather than CAD-to-image.

\paragraph{Shape Retrieval Challenges and RGB-D Datasets}

Shape retrieval challenges have recently been organized as part of the Eurographics 3DOR~\cite{huashrec2017,phamshrec2018}. 
Here, the task was formulated as matching of object instances from ScanNet~\cite{dai2017scannet} and SceneNN~\cite{hua2016scenenn} to CAD models from the ShapeNetSem dataset~\cite{shapenet2015}. 
Evaluation  only considered binary in-category vs out-of-category (and sub-category) match as the notion of relevance. As such, this evaluation does not address the alignment quality between scan objects and CAD models, which is our focus.

ScanNet~\cite{dai2017scannet} provides aligned CAD models for a small subset of the annotated object instances (for only 200 objects out of the total 36000).
Moreover, the alignment quality is low with many object category mismatches and alignment errors, as the annotation task was performed by crowdsourcing.
The PASCAL 3D+~\cite{xiang2014beyond} dataset annotates 13898 objects in the PASCAL VOC images with coarse 3D poses defined against representative CAD models.
ObjectNet3D~\cite{xiang2016objectnet3d} provides a dataset of CAD models aligned to 2D images, approximately 200K object instances in 90K images.
The IKEA objects~\cite{lim2013parsing} and Pix3D~\cite{sun2018pix3d} datasets similarly provide alignments of a small set of identifiable CAD models to 2D images of the same objects in the real world; the former has 759 images annotated with 90 models, the latter has 10069 annotated with 395 models.

No existing dataset provides fine-grained object instance alignments at the scale of our Scan2CAD dataset with \OBJECTS{} CAD models (\MODELS{} unique instances) annotated to their scan counterpart distributed on \SCANS{} 3D scans.

\section{Overview}
\paragraph{Task}
We address alignment between clean CAD models and noisy, incomplete 3D scans from RGB-D fusion, as illustrated in \autoref{fig:teaser}.
Given a 3D scene $\mathbb{S}$ and a set of 3D CAD models $\mathbb{M}=\{m_i\}$, the goal is to find a 9DoF transformation $T_i$ (3 degrees for translation, rotation, and scale each) for every CAD model $m_i$ such that it aligns with a semantically matching object $\mathbb{O} = \{ o_j\}$ in the scan.
One important note is that we cannot guarantee the existence of 3D models which exactly matches the geometry of the scan objects.

\vspace{-1em}
\paragraph{Dataset and Benchmark}
In \autoref{sec:dataset}, we introduce the construction of our \OURS{} dataset. 
We propose an annotation pipeline designed for use by trained annotators.
An annotator first inspects a 3D scan and selects a model from a CAD database that is geometrically similar to a target object in the scan.
Then, for each model, the annotator defines corresponding keypoint pairs between the model and the object in the scan.
From these keypoints, we compute ground truth 9DoF alignments.
We annotate the entire ScanNet dataset and use the original training, validation, and test splits to establish our alignment benchmark. 

\vspace{-1em}
\paragraph{Heatmap Prediction Network}
In \autoref{sec:network}, we propose a 3D CNN taking as input a volume around a candidate keypoint in a scan and a volumetric representation of a CAD model.
The network is trained to predict a correspondence heatmap over the CAD volume, representing the likelihood that the input keypoint in the scan is matching with each voxel.
The heatmap prediction is formulated as a classification problem, which is easier to train than regression, and produces sparse correspondences needed for pose optimization.

\vspace{-1em}
\paragraph{Alignment Optimization}
\autoref{sec:alignment} describes our variational alignment optimization.
To generate candidate correspondence points in the 3D scan, we detect Harris keypoints, and predict correspondence heatmaps for each Harris keypoint and CAD model. Using the predicted heatmaps we find optimal 9DoF transformations. False alignments are pruned via a geometric confidence metric.

\section{Dataset}
\label{sec:dataset}

Our \OURS{} dataset builds upon the 3D scans from ScanNet~\cite{dai2017scannet} and CAD models from ShapeNet~\cite{shapenet2015}.
Each \textit{scene} $\mathbb{S}$ contains multiple \textit{objects} $\mathbb{O}=\{o_i\}$, where each \textit{object} $o_i$ is matched with a ShapeNet CAD model $m_i$ and both share multiple keypoint pairs (correspondences) and one transformation matrix $T_i$ defining the alignment.
Note that ShapeNet CAD models have a consistently defined front and upright orientation which induces an amodal tight oriented bounding box for each scan object, see \autoref{fig:obb}.

\subsection{Data Annotation}
\begin{figure}
\begin{center}
    \begin{subfigure}{\columnwidth}
         \includegraphics[width=\linewidth]{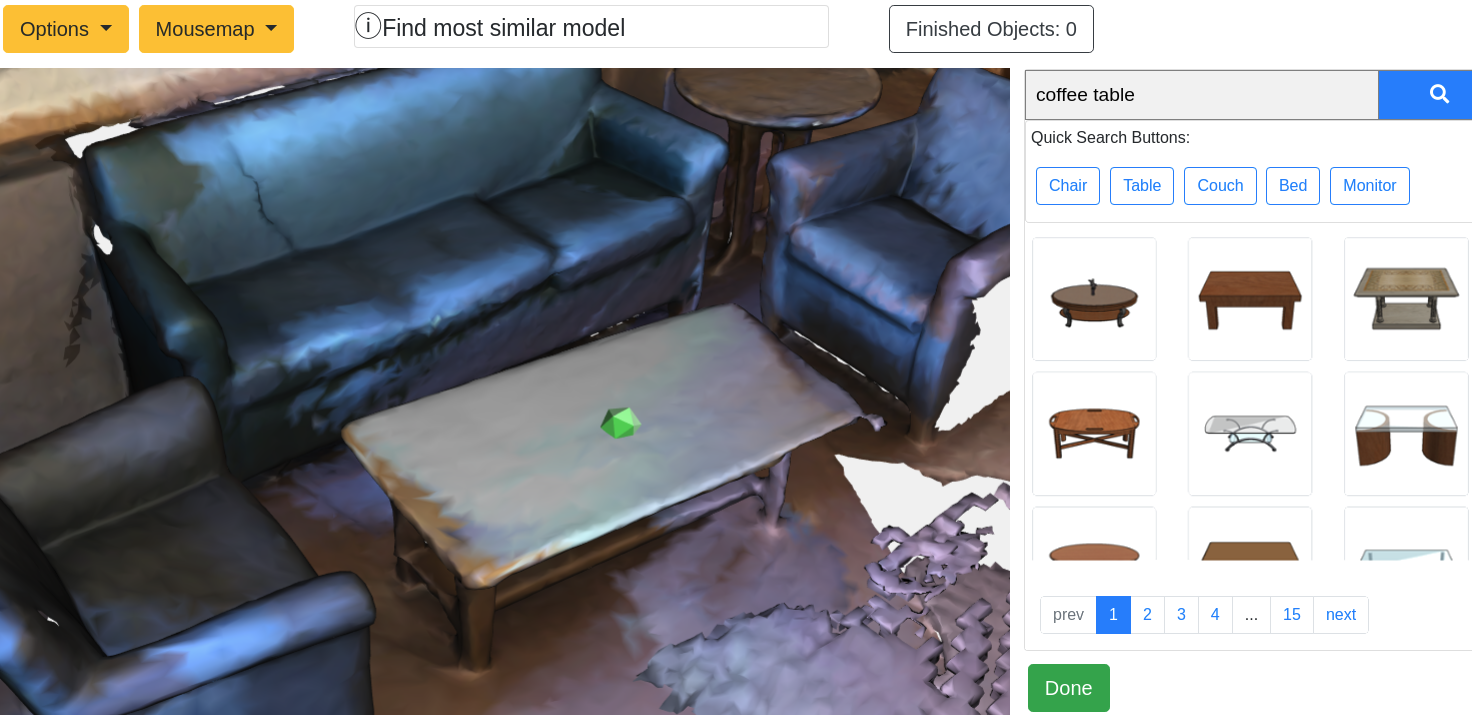}
        \caption{First step: Retrieval view.}
        \label{fig:webapp_retrieval}
    \end{subfigure}    
    \begin{subfigure}{\columnwidth}
         \includegraphics[width=\linewidth]{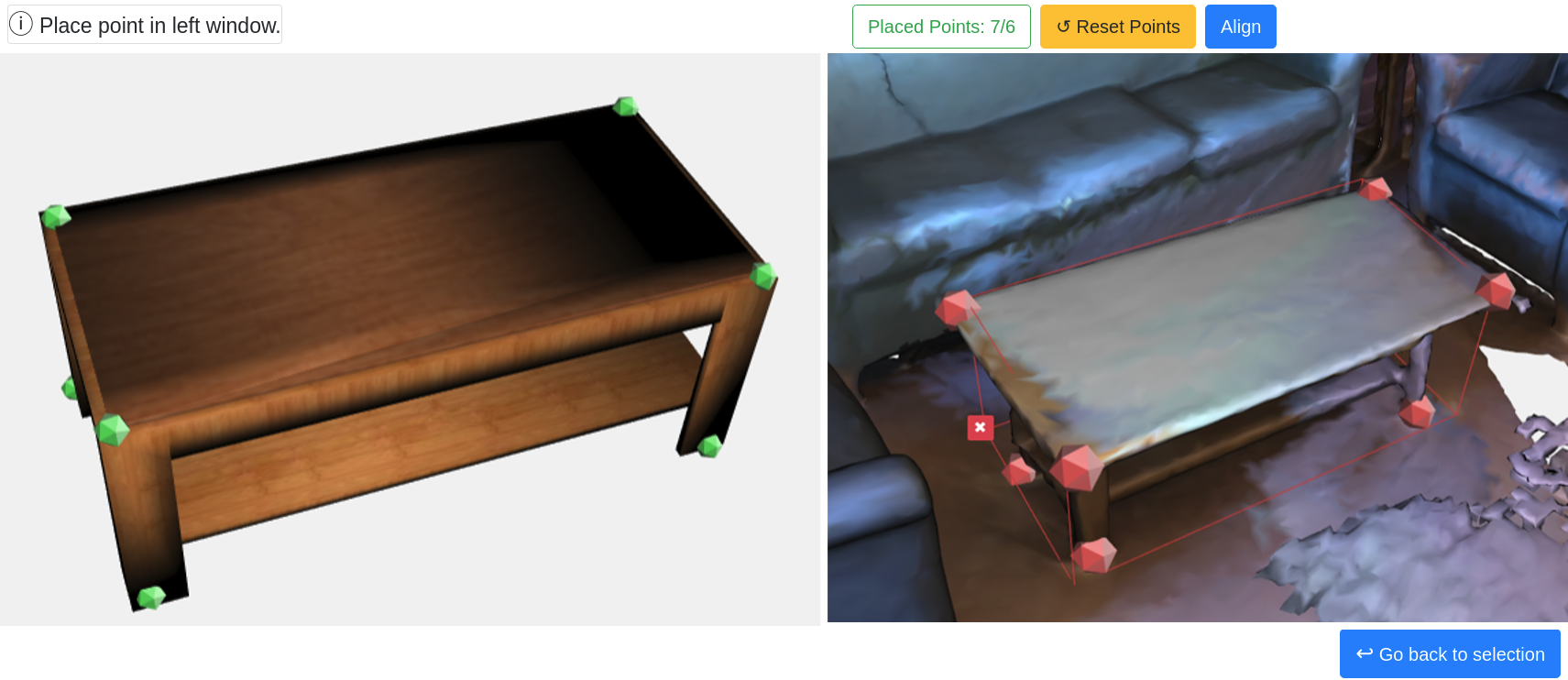}
        \caption{Second step: Alignment view.}
        \label{fig:webapp_alignment}
    \end{subfigure}
\end{center}
\vspace{-0.5cm}
\caption{Our annotation web interface is a two-step process. (a) After the user places an anchor on the scan surface, class-matching CAD models are displayed on the right. (b) Then the user annotates keypoint pairs between the scan and CAD model from which we derive the ground truth 9DoF transformation.}
\label{fig:webapp}
\end{figure}

\begin{figure}[htb!]
\begin{center}
    \begin{subfigure}[h]{0.47\columnwidth}
         \includegraphics[width=\linewidth]{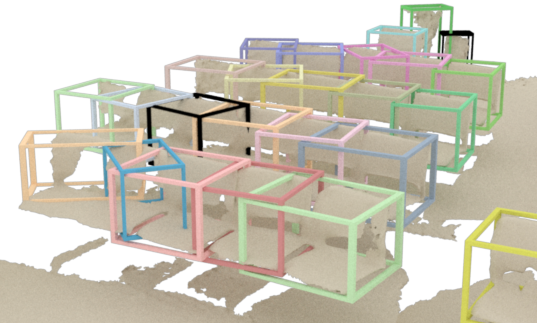}
        \label{fig:bbox_seg0}
    \end{subfigure}    
    \begin{subfigure}[h]{0.47\columnwidth}
         \includegraphics[width=\linewidth]{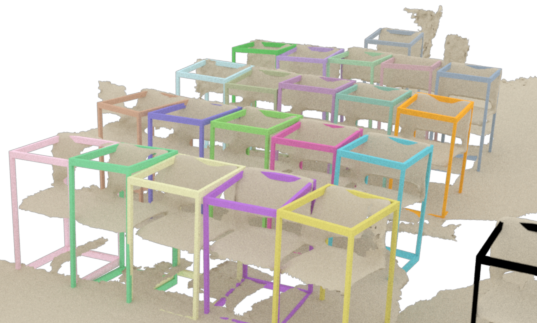}
        \label{fig:bbox_ann0}
    \end{subfigure}
\end{center}
\vspace{-0.9cm}
\caption{(Left) Oriented bounding boxes (OBBs) computed from the instance segmentation of ScanNet~\cite{dai2017scannet} are often incomplete due to missing geometry (e.g., in this case, missing chair legs). (Right) Our OBBs are derived from the aligned CAD models and are thus complete.
    \vspace{-0.4cm}
    }
\label{fig:obb}
\end{figure}

The annotation is done via a web application that allows for simple scaling and distribution of annotation jobs; see \autoref{fig:webapp}.
The annotation process is separated into two steps.
The first step is object \textit{retrieval}, where the user clicks on a point on the 3D scan surface, implicitly determining an object category label from the ScanNet object instance annotations.
We use the instance category label as query text in the ShapeNet database to retrieve and display all matching CAD models in a separate window as illustrated in \autoref{fig:webapp_retrieval}. After selecting a CAD model the user performs \textit{alignment}.

In the alignment step, the user sees two separate windows in which the CAD model (left) and the scan object (right) are shown (see \autoref{fig:webapp_alignment}).
Keypoint correspondences are defined by alternately clicking paired points on the CAD model and scan object.
We require users to specify at least 6 keypoint pairs to determine a robust ground truth transformation.
After keypoint pairs are specified, the alignment computation is triggered by clicking a button.
This alignment (given exact 1-to-1 correspondences) is solved with the genetic algorithm \textit {CMA-ES}~\cite{hansen2003reducing,hansen2009benchmarking} that minimizes the point-to-point distance over 9 parameters.
In comparison to gradient-based methods or Procrustes superimposition method, we found this approach to perform significantly better in reliably returning high-quality alignments regardless of initialization.

The quality of these keypoint pairs and alignments was verified in several verification passes, with re-annotations performed to ensure a high quality of the dataset. The verification passes were conducted by the authors of this work.

A subset of the ShapeNet CAD models have symmetries that play an important role in making correspondences. Hence, we annotated all ShapeNet CAD models used in our dataset with their rotational symmetries to prevent false negatives in evaluations. We defined 2-fold ($C_2$), 4-fold ($C_4$) and infinite ($C_{\infty}$) rotational symmetries around a canonical axis of the object.

\subsection{Dataset Statistics}
\label{sec:dataset_stats}

The annotation process yielded \KEYPOINTS{} keypoint pairs on \OBJECTS{} (\MODELS{} unique) CAD models with their respective scan counterpart distributed on a total of \SCANS{}. Approximately $28\%$ out of the \MODELS{} CAD models have a symmetry tag (either $C_2$, $C_4$ or $C_\infty$). 

Given the complexity of the task and to ensure high quality annotations, we employed 7 part-time annotators (in contrast to crowd-sourcing). On average, each scene has been edited $1.76$ times throughout the re-annotation cycles. The top 3 annotated model classes are chairs, tables and cabinets which arises due to the nature of indoor scenes in ScanNet. The number of objects aligned per scene ranges from $1$ to $40$ with an average of $9.3$. It took annotators on average of $2.48 \text{min}$ to align each object, where the time to find an appropriate CAD model dominated the time for keypoint placement. The average annotation time for an entire scene is $20.52 \text{min}$.

It is interesting to note that manually placed keypoint correspondences between scans and CAD models differ significantly from those extracted from a Harris corner detector. 
Here, we compare the mean distance from the annotated CAD keypoint to: (1) the corresponding annotated scan keypoint ($=3.5cm$) and (2) the nearest Harris keypoint in the scan ($=12.8cm$). 

\subsection{Benchmark}
Using our annotated dataset, we designed a benchmark to evaluate scan-to-CAD alignment methods.
A model alignment is considered successful only if the category of the CAD model matches that of the scan object \emph{and} the pose error is within translation, rotational, and scale bounds relative to the ground truth CAD.
We do not enforce strict instance matching (i.e., matching the exact CAD model of the ground truth annotation) as ShapeNet models typically do not identically match real-world scanned objects. 
Instead, we treat CAD models of the same category as interchangeable (according to the ShapeNetCorev2 \textit{top-level synset}).

Once a CAD model is determined to be aligned correctly, the ground truth counterpart is removed from the candidate pool in order to prevent multiple alignments to the same object. 
Alignments are fully parameterized by 9 pose parameters. A quantitative measure based on bounding box overlap (IoU) can be readily calculated with these parameters as CAD models are defined on the unit box. The error thresholds for a successful alignment are set to $\epsilon_t \leq 20\text{cm}$, $\epsilon_r \leq 20^\circ$, and $\epsilon_s \leq 20\%$ for translation, rotation, and scale respectively (for extensive error analysis please see the supplemental). The rotation error calculation takes $C_2$, $C_4$ and $C_\infty$ rotated versions into account.

The \OURS{} dataset and associated symmetry annotations is available to the community. 
For standardized comparison of future approaches, we operate an automated test script on a hidden test set. 

\section{Correspondence Prediction Network}
\label{sec:network}

\subsection{Data Representation}
Scan data is represented by its signed distance field (SDF) encoded in a volumetric grid and generated through \textit{volumetric fusion}~\cite{curless1996volumetric} from the depth maps of the RGB-D reconstruction (voxel resolution = $3cm$, truncation = $15cm$).
For the CAD models, we compute unsigned distance fields (DF) using the level-set generation toolkit by Batty~\cite{battysdf}. 

\subsection{Network Architecture} \label{subsec:network_architecture}
\begin{figure*}[tp]
\begin{center}
   \includegraphics[width=0.85\linewidth]{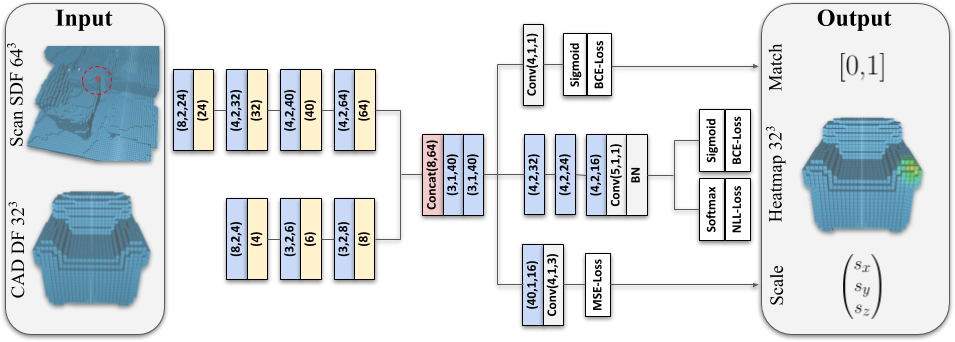}
    \end{center}
    \vspace{-0.4cm}
   \caption{3D CNN architecture of our \OURS{} approach: we take as input SDF chunks around a given keypoint from a 3D scan and the DF of a CAD model. 
   These are encoded with 3D CNNs to learn a shared embedding between the synthetic and real data; from this, we classify whether there is semantic compatibility between both inputs (top), predict a correspondence heatmap in the CAD space (middle) and the scale difference between the inputs (bottom).
    \vspace{-0.4cm}
    }
\label{fig:architecture}
\end{figure*}

Our architecture takes as input a pair of voxel grids: A SDF centered at a point in the scan with a large receptive field at $64^3$ size, and a DF of a particular CAD model at $32^3$ size. We use a series of convolutional layers to separately encode each input stream (see \autoref{fig:architecture}).
The two encoders compress the volumetric representation into compact feature volumes of $4^3 \times 64$ (scan) and $4^3 \times 8$ (CAD) which are then concatenated before passing to the decoder stage. The decoder stage predicts three output targets, heatmap, compatibility, and scale, described as follows:

\paragraph{Heatmap}
The first output is a heatmap $H: \Omega \rightarrow [0,1]$ over the $32^3$ voxel domain $\Omega \in \mathbb{N}^3$ of the CAD model producing the voxel-wise correspondence probability.
This indicates the probability of matching each voxel in $\Omega$ to the center point of the scan SDF. We train our network using a combined binary cross-entropy (BCE) loss and a negative log-likelihood (NLL) to predict the final heatmap $H$. The raw output $S: \Omega \rightarrow \mathbb{R}$ of the last layer in the decoder is used to generate the heatmaps:
\begin{align}
&H_1: \Omega \rightarrow [0,1], \quad x \mapsto \text{sigmoid}(S(x)) \notag \\
&H_2: \Omega \rightarrow [0,1], \quad x \mapsto \text{softmax}(S(x)) \notag \\
&\mathcal{L}_{H} =\sum_{x \in \Omega} w(x)\cdot \text{BCE}(H_1, H_\text{GT}) + \sum_{x \in \Omega} v\cdot\text{NLL}(H_2, H_\text{GT}) \notag
\end{align}
where $w(x) = 64.0 \text{ if } x > 0.0 \text{ else } 1.0, v=64$ are weighting factors to increase the signal of the few sparse positive keypoint voxels in the voxel grid ($\approx 99\%$ of the target voxels have a value equal to $0$).
The combination of the sigmoid and softmax terms is a compromise between high recall but low precision using sigmoid, and more locally sharp keypoint predictions using softmax over all voxels. The final target heatmap, used later for alignment, is constructed with an element-wise multiplication of both heatmap variations: $H = H_1 \circ H_2$.

\paragraph{Compatibility} The second prediction target is a single probability score $\in [0,1]$ indicating semantic compatibility between scan and CAD.
This category equivalence score is $0$ when the category labels are different (e.g., scan table and CAD chair) and $1$ when the category labels match (e.g., scan chair and CAD chair). The loss function for this output is a sigmoid function followed by a BCE loss:
\begin{align}
    &\mathcal{L}_\text{compat.} = \text{BCE}(\text{sigmoid}(x), x_\text{GT})  \notag 
\end{align}
\paragraph{Scale} The third output predicts the scale $\in \mathbb{R}^3$ of the CAD model to the respective scan. Note that we do not explicitly enforce positivity of the predictions. This loss term is a mean-squared-error (MSE) for a prediction $x \in \mathbb{R}^3$:
\begin{align}
    &\mathcal{L}_\text{scale} = \text{MSE}(x, x_\text{GT}) = \Vert x - x_\text{GT} \Vert_2^2  \notag 
\end{align}

Finally, to train our network, we use a weighted combination of the presented losses:
\begin{align}
    \mathcal{L} = 1.0\mathcal{L}_\text{H} + 0.1 \mathcal{L}_\text{compat.} + 0.2 \mathcal{L}_\text{scale} \notag 
\end{align}
where the weighting of each loss component was empirically determined for balanced convergence.

\subsection{Training Data Generation}
\label{sec:data_generation}
\paragraph{Voxel Grids} Centered scan volumes are generated by projecting the annotated keypoint into the scan voxel grid and then cropping around it with  a crop window of $63^3$.
Ground truth heatmaps are generated by projecting annotated keypoints (and any symmetry-equivalent keypoints) into the CAD voxel grid. We then use a Gaussian blurring kernel ($\sigma = 2.0$) on the voxel grid to account for small keypoint annotation errors and to avoid sparsity in the loss residuals.

\paragraph{Training Samples} With our annotated dataset we generate $N_{P, \text{ann.}} = \KEYPOINTS{} $ positive training pairs where one pair consists of an annotated scan keypoint and the corresponding CAD model. Additionally, we create $N_{P,\text{aug.}} = 10\cdot N_{P, \text{ann.}}$, augmented positive keypoint pairs by randomly sampling points on the CAD surface, projecting them to the scan via the ground truth transformation and rejecting if the distance to the surface in the scan $\geq 3cm$. In total we generate $N_P = N_{P, \text{ann.}} + N_{P,\text{aug.}}$ positive training pairs.

Negative pairs are generated in two ways:
(1) Randomly choosing a voxel point in the scan and a random CAD model (likelihood of false negative is exceedingly low).
(2) Taking an annotated scan keypoint and pairing it with a random CAD model of different class. 
We generate $N_N = N_P$ negative samples with (1) and $N_{HN} = N_P$ with (2).

Hence, the training set has a positives-to-negatives ratio of 1:2 ($N_P$ : $N_N + N_{HN}$). We found an over-representation of negative pairs gives satisfactory performance on the compatibility prediction.

\subsection{Training Process}
\label{sec:training}
We use an SGD optimizer with a batch size of 32 and an initial learning rate of $0.01$, which is decreased by $1/2$ every $50$K iterations. We train for 250K iterations ($\approx 62.5$ hours). The weights are initialized randomly. The losses of the heatmap prediction stream and the scale prediction stream are masked such that only positive samples make up the residuals for back-propagation. 

The CAD encoder is pre-trained with an auto-encoder on ShapeNet models with a reconstruction task and a $MSE$ as loss function. All models of ShapeNetCore ($\approx 55K$) are used for pre-training and the input and output dimensions are $32^3$ distance field grids. The network is trained with SGD until convergence ($\approx 50$ epochs).

\section{Alignment Optimization}
\label{sec:alignment}
\paragraph{Filtering}
The input to our alignment optimization is a representative set of Harris keypoints $\mathbb{K} = \{ p_j \} , \ j=1 \dots N_0$ from a scene $\mathbb{S}$ and a set of CAD models $\mathbb{M}=\{m_i\}$.
The correspondences between $\mathbb{K}$ and $\mathbb{M}$ were established by the correspondence prediction from the previous stage (see \autoref{sec:network}) where each keypoint $p_j$ is tested against every model $m_i$.

Since not every keypoint $p_j$ semantically matches to every CAD model $m_i$, we reject correspondences based on the compatibility prediction of our network. The threshold for rejecting $p_j$ is determined by the Otsu thresholding scheme~\cite{otsu1979threshold}.
In practice this method turned out to be much more effective than a fixed threshold. After the filtering there are $N \leq N_0$ (usually $N \approx 0.1N_0$) correspondence pairs to be used for the alignment optimization.

\paragraph{Variational Optimization}
From the remaining $\mathbb{K}_\text{filter.} \subset{\mathbb{K}}$ Harris keypoints, we construct \textit{point-heatmap} pairs $(p_j, H_j)$ for each CAD model $m_i$, with $p_j \in \mathbb{R}^3$ a point in the scan and $H_j: \Omega \rightarrow [0,1]$ a heatmap.

In order to find an optimal pose we construct the following minimization problem:
\begin{align}
    \text{c}_\text{vox} &= T_{\text{world} \rightarrow \text{vox}} \cdot T_{m_i}(a, s) \cdot p_j \notag \\
    f &= \min_{a ,s}\quad \sum_j^N (1 - H_j(\text{c}_\text{vox}))^2 + \lambda_s \Vert s \Vert_2^2 \label{eq:optimization}
\end{align}
where $\text{c}_\text{vox}$ is a voxel coordinate, $T_{\text{world} \rightarrow \text{vox}}$ denotes a transformation that maps world points into the voxel grid for look-ups, $a$ denotes the coordinates of the Lie algebra (for rotation and translation), $s$ defines the scale, and $\lambda_s$ defines the scale regularization strength. $a,s$ compose a transformation matrix $T_{m_i} = \psi(a_{m_i}, s_{m_i})$:
\begin{align}
     \psi : ~& \mathbb{R}^6 \times \mathbb{R}^3 \rightarrow \mathbb{R}^{4 \times 4}, \notag \\
     &a,s \mapsto \text{expm}\left( \begin{bmatrix} \Gamma(a_{1,2,3}) & a_{4,5,6} \\ 0 & 0 \end{bmatrix} \right) \cdot \begin{bmatrix} s & 0 \\ 0 & 1\end{bmatrix} \notag
\end{align}
where $\Gamma$ is the hat map, $\text{expm}$ is the matrix exponential.

We solve \autoref{eq:optimization} using the Levenberg-Marquardt (LM) algorithm.
As we can suffer from zero-gradients (especially at bad initialization), we construct a scale-pyramid from the heatmaps which we solve in coarse-to-fine fashion.

In each LM step we optimize over the incremental change and update the parameters as following: $T_{m_i}^{k + 1}  \leftarrow \phi(a^*, s^*) \cdot T_{m_i}^k$ where $a^*, s^*$ are the optimal parameters.
As seen in \autoref{eq:optimization}, we add a regularization on the scale in order to prevent degenerate solutions which can appear for very large scales.

By restarting the optimization with different translation parameters (i.e., varying initializations), we obtain multiple alignments per CAD model $m_i$. We then generate as many CAD model alignments as required for a given scene in the evaluation. Note, in a ground truth scene one unique CAD model $m_i$ can appear in multiple locations e.g., chairs in conference rooms.

\paragraph{Pruning}
Finally, there will be alignments of various CAD models into a scene where a subset will be misaligned. In order to select only the best alignments and prune potential misalignments we use a confidence metric similar to~\cite{li2015database}; for more detail, we refer to the  appendix.

\begin{table*}[tp]
\begin{center}
\footnotesize
\begin{tabular}{|l|rrrrrrrrr|r|r|}
\hline   
base [+variations, ...]                       & bath             & bookshelf        & cabinet          & chair            & display          & sofa             & table            & trash bin        & other            & class avg.       & avg.           \\ \hline
\hline
+sym & 46.88          & 44.39          & 40.49          & 64.46          & 26.85          & 56.26          & 47.15          & 38.43          & 24.68          & 43.29          & 48.01          \\
+sym,+scale           & 51.35          & 45.46          & 45.24          & 66.94          & 29.88          & 64.78          & 48.30          & 38.00          & 28.65          & 46.51          & 50.85          \\
+sym,+CP        & 59.32          & 51.93          & 55.11          & 70.99          & 41.58          & 66.77          & 53.74          & 43.39          & 42.93          & 53.97          & 60.44          \\
+scale,+CP         & 45.24          & 45.85          & 47.16          & 61.55          & 27.65          & 51.92          & 41.21          & 31.13          & 29.62          & 42.37          & 47.64 \\ 
+sym,+scale,+CP      & 56.05          & 51.28          & \textbf{57.45} & 72.64          & 36.36          & 70.63          & 52.28          & 46.80          & 43.32          & 54.09          & 60.43          \\
+sym,+scale,+CP,+PT (3/3 fix)     & 57.03          & 50.63          & 56.76          & 70.39          & 39.74          & 65.00          & 52.03          & \textbf{46.87}          & 41.83            & 53.36          & 58.61 \\
+sym,+scale,+CP,+PT  (1/3 fix)     & \textbf{60.08} & \textbf{58.62} & 56.35          & \textbf{73.92} & \textbf{44.19} & \textbf{75.08} & \textbf{56.80} & 45.78 & \textbf{46.53}          & \textbf{57.48} & \textbf{63.94} \\ \hline   
\end{tabular}
\end{center}
\vspace{-0.6cm}
\caption{Correspondence prediction F1-scores in $\%$ for variations of our correspondence prediction network. 
We evaluate the effect of symmetry (sym), predicting scale (scale), predicting compatibility (CP), encoder pre-training (PT), and pre-training with parts of the encoder fixed (\#fix), see \autoref{sec:network} for more detail regarding our network design and training scheme.
}
\label{tab:correspondence}
\end{table*}
\begin{figure*}[bp]
\begin{center}
\includegraphics[width=0.87\linewidth]{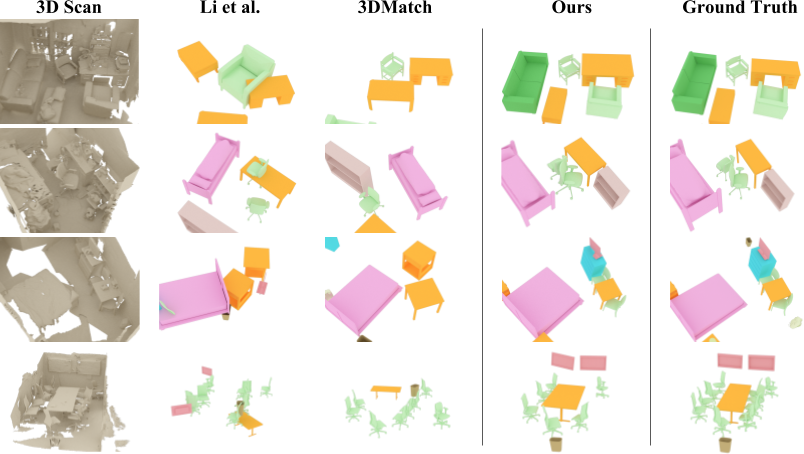}
\end{center}
\vspace{-0.6cm}
\caption{Qualitative comparison of alignments on four different test ScanNet~\cite{dai2017scannet} scenes. 
Our approach to learning geometric features between real and synthetic data produce much more reliable keypoint correspondences, which coupled with our alignment optimization, produces significantly more accurate alignments.
\vspace{-0.5cm}
}
\label{fig:results}
\end{figure*}

\section{Results}

\subsection{Correspondence Prediction}

To quantify the performance of correspondence heatmap predictions, we evaluate the voxel-wise F1-score for a prediction and its Gaussian-blurred target. The task is challenging and by design $\frac{2}{3}$ test samples are false correspondences, $\approx 99\%$ of the target voxels are 0-valued, and only a single 1-valued voxel out of $32^3$ voxels exists. The F1-score will increase only by identifying true correspondences. As seen in \autoref{tab:correspondence}, our best 3D CNN achieves $63.94\%$.

\autoref{tab:correspondence} additionally addressed our design choices; in particular, we evaluate the effect of using pre-training (PT), using compatibility (CP) as a proxy loss (defined in \autoref{subsec:network_architecture}), enabling symmetry awareness (sym), and predicting scale  (scale).
Here, a pre-trained network reduces overfitting, enhancing generalization capability.
Optimizing for compatibility strongly improves heatmap prediction as it efficiently detects false correspondences.
While predicting scale only slightly influences the heatmap predictions, it becomes very effective for the later alignment stage.
Additionally, incorporating symmetry enables significant improvement by explicitly disambiguating symmetric keypoint matches.

\subsection{Alignment}
In the following, we compare our approach to other handcrafted feature descriptors:  FPFH~\cite{rusu2009fast}, SHOT~\cite{tombari2010signature}, Li et al.~\cite{li2015database} and a learned feature descriptor: 3DMatch~\cite{zeng20173dmatch} (trained on our \OURS{} dataset).  
We combine these descriptors with a RANSAC outlier rejection method to obtain pose estimations for an input set of CAD models. 
A detailed description of the baselines can be found in the appendix.
As seen in \autoref{tab:alignment}, our best method achieves $31.68\%$ and outperforms all other methods by a significant margin.
We additionally show qualitative results in \autoref{fig:results}.
Compared to state-of-the-art handcrafted feature descriptors, our learned approach powered by our \OURS{} dataset produces considerably more reliable correspondences and CAD model alignments.
Even compared to the learned descriptor approach of 3DMatch, our explicit learning across the synthetic and real domains coupled with our alignment optimization produces notably improved CAD model alignment.

\begin{figure}[htb!]
\begin{center}
\includegraphics[width=0.9\linewidth]{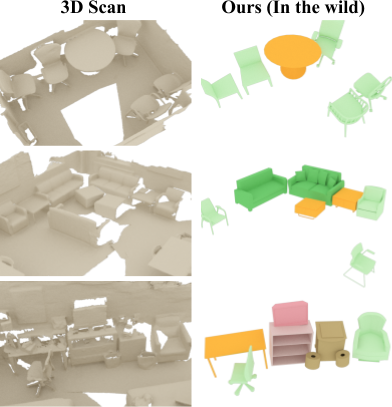}
\end{center}
\vspace{-0.6cm}
\caption{Unconstrained scenario where instead of having a ground truth set of CAD models given, we use a set of 400 randomly selected CAD models from ShapeNetCore~\cite{shapenet2015}, more closely mimicking a real-world application scenario.
\vspace{-0.6cm}
}
\label{fig:unconstrained}
\end{figure}

\autoref{fig:unconstrained} shows the capability of our method to align in an unconstrained real-world setting where ground truth CAD models are not given, we instead provide a set of 400 random CAD models from ShapeNet~\cite{shapenet2015}.

\begin{table*}[htb!]
\begin{center}
\footnotesize
\begin{tabular}{|l|r r r r r r r r r | r | r|}
\hline
   & bath             & bookshelf        & cabinet          & chair            & display          & sofa             & table            & trash bin        & other            & class avg.       & avg.           \\ \hline
\hline
FPFH (Rusu et al.~\cite{rusu2009fast}) & 0.00 & 1.92 & 0.00 & 10.00 & 0.00 & 5.41 & 2.04 & 1.75 & 2.00 & 2.57 & 4.45 \\ 
SHOT (Tombari et al.~\cite{tombari2010signature}) & 0.00 & 1.43 & 1.16 & 7.08 & 0.59 & 3.57 & 1.47 & 0.44 & 0.75 & 1.83 & 3.14 \\
Li et al.~\cite{li2015database} & 0.85 & 0.95 & 1.17 & 14.08 & 0.59 & 6.25 & 2.95 & 1.32 & 1.50 & 3.30 & 6.03 \\ 
3DMatch (Zeng et al.~\cite{zeng20173dmatch}) & 0.00 & 5.67 & 2.86 & 21.25 & 2.41 & 10.91 & 6.98 & 3.62 & 4.65 & 6.48 & 10.29  \\ \hline
Ours: +sym & 24.30          & 10.61          & 5.97           & 9.49           & 3.90           & 25.26          & 12.34          & 10.74          & 3.58           & 11.80          & 8.772          \\ 
Ours: +sym,+scale           & 18.99          & 13.61          & 7.24           & 14.73          & 9.76           & 41.05          & 14.04          & 5.26           & 6.29           & 14.55          & 11.48          \\ 
Ours: +sym,+CP        & 35.90          & 32.35          & 28.64          & 40.48          & 18.85          & 60.00          & 33.11          & 28.42          & 16.89          & 32.74          & 29.42          \\
Ours: +scale,+CP          & 34.18          & 31.76          & 21.82          & 37.02          & 14.75          & 50.53          & 32.31          & \textbf{31.05} & 11.59          & 29.45          & 26.75          \\ 
Ours: +sym,+scale,+CP                  & 36.20          & \textbf{36.40} & \textbf{34.00} & \textbf{44.26} & 17.89          & \textbf{70.63} & 30.66          & 30.11          & 20.60          & \textbf{35.64} & \textbf{31.68} \\ 
Ours: +sym,+scale,+CP,+PT (3/3 fix)     & \textbf{37.97} & 30.15          & 28.64          & 41.55          & 19.51          & 57.89          & 33.85          & 20.00          & 17.22          & 31.86          & 29.27          \\
Ours: +sym,+scale,+CP,+PT (1/3 fix)     & 34.81          & \textbf{36.40} & 29.00          & 40.60          & \textbf{23.25} & 66.00          & \textbf{37.64} & 24.32          & \textbf{22.81} & 34.98          & 31.22          \\ \hline
\end{tabular}
\end{center}
\vspace{-0.5cm}
\caption{Accuracy comparison ($\%$) on our CAD alignment benchmark.
While handcrafted feature descriptors can achieve some alignment on more featureful objects (e.g., chairs, sofas), they do not tolerate well the geometric discrepancies between scan and CAD data -- which remains difficult for the learned keypoint descriptors of 3DMatch.
\OURS{} directly addresses this problem of learning features that generalize across these domains, thus significantly outperforming state of the art.
}
\label{tab:alignment}
\vspace{-0.5cm}
\end{table*}

\section{Limitations}
While the focus of this work is mainly on the alignment between 3D scans and CAD models, we only provide a basic algorithmic component for retrieval (finding the most similar model). This necessitates an exhaustive search over a set of CAD models. 
We believe that one of the immediate next steps in this regard would be designing a neural network architecture that is specifically trained on shape similarity between scan and CAD geometry to introduce more efficient CAD model retrieval. 
Additionally, we currently only consider geometric information, and it would also be intresting to introduce learned color features into the correspondence prediction, as RGB data is typically higher-resolution than depth or geometry, and could potentially improve alignment results.

\section{Conclusion}

In this work, we presented \OURS{}, which aligns a set of CAD models to 3D scans by predicting correspondences in form of heatmaps and then optimizes over these correspondence predictions.
First, we introduce a new dataset of 9DoF CAD-to-scan alignments with \KEYPOINTS{} pairwise keypoint annotations defining the alignment of \OBJECTS{} objects. Based on this new dataset, we design  a 3D CNN to predict correspondence heatmaps between a CAD model and a 3D scan. From these predicted heatmaps, we formulate a variational energy minimization that then finds the optimal 9DoF pose alignments between CAD models and the scan, enabling effective transformation of noisy, incomplete RGB-D scans into a clean, complete CAD model representation.
This enables us to achieve significantly more accurate results than state-of-the-art approaches, and we hope that our dataset and benchmark will inspire future work towards bringing RGB-D scans to CAD or artist-modeled quality.

\section*{Acknowledgements}
We would like to thank the expert annotators 
Soh Yee Lee, 
Rinu Shaji Mariam,
Suzana Spasova,
Emre Taha, 
Sebastian Thekkekara, and
Weile Weng
for their efforts in building the Scan2CAD dataset.
This work is supported by Occipital, the ERC Starting Grant Scan2CAD (804724), and a Google Faculty Award.
We would also like to thank the support of the TUM-IAS, funded by the German Excellence Initiative and the European Union Seventh Framework Programme under grant agreement n° 291763, for the TUM-IAS Rudolf M{\"o}{\ss}bauer Fellowship and Hans-Fisher Fellowship (Focus Group Visual Computing).

{\small
\bibliographystyle{ieee}
\bibliography{egbib}

\begin{thebibliography}{10}\itemsep=-1pt

\bibitem{battysdf}
C.~Batty.
\newblock {SDFGen}.
\newblock \url{https://github.com/christopherbatty/SDFGen}.

\bibitem{Matterport3D}
A.~Chang, A.~Dai, T.~Funkhouser, M.~Halber, M.~Niessner, M.~Savva, S.~Song,
  A.~Zeng, and Y.~Zhang.
\newblock {Matterport3D}: Learning from {RGB-D} data in indoor environments.
\newblock {\em International Conference on 3D Vision (3DV)}, 2017.

\bibitem{shapenet2015}
A.~X. Chang, T.~Funkhouser, L.~Guibas, P.~Hanrahan, Q.~Huang, Z.~Li,
  S.~Savarese, M.~Savva, S.~Song, H.~Su, J.~Xiao, L.~Yi, and F.~Yu.
\newblock {ShapeNet: An Information-Rich 3D Model Repository}.
\newblock Technical Report arXiv:1512.03012 [cs.GR], Stanford University ---
  Princeton University --- Toyota Technological Institute at Chicago, 2015.

\bibitem{chen2013scalable}
J.~Chen, D.~Bautembach, and S.~Izadi.
\newblock Scalable real-time volumetric surface reconstruction.
\newblock {\em ACM Transactions on Graphics (TOG)}, 32(4):113, 2013.

\bibitem{choi2015robust}
S.~Choi, Q.-Y. Zhou, and V.~Koltun.
\newblock Robust reconstruction of indoor scenes.
\newblock In {\em 2015 IEEE Conference on Computer Vision and Pattern
  Recognition (CVPR)}, pages 5556--5565. IEEE, 2015.

\bibitem{curless1996volumetric}
B.~Curless and M.~Levoy.
\newblock A volumetric method for building complex models from range images.
\newblock In {\em Proceedings of the 23rd annual conference on Computer
  graphics and interactive techniques}, pages 303--312. ACM, 1996.

\bibitem{dai2017scannet}
A.~Dai, A.~X. Chang, M.~Savva, M.~Halber, T.~Funkhouser, and M.~Nie{\ss}ner.
\newblock {ScanNet}: Richly-annotated {3D} reconstructions of indoor scenes.
\newblock In {\em Proc. Computer Vision and Pattern Recognition (CVPR), IEEE},
  2017.

\bibitem{dai2017bundlefusion}
A.~Dai, M.~Nie{\ss}ner, M.~Zollh{\"o}fer, S.~Izadi, and C.~Theobalt.
\newblock Bundlefusion: Real-time globally consistent 3d reconstruction using
  on-the-fly surface reintegration.
\newblock {\em ACM Transactions on Graphics (TOG)}, 36(3):24, 2017.

\bibitem{dai2017complete}
A.~Dai, C.~R. Qi, and M.~Nie{\ss}ner.
\newblock Shape completion using 3d-encoder-predictor cnns and shape synthesis.
\newblock In {\em Proc. Computer Vision and Pattern Recognition (CVPR), IEEE},
  2017.

\bibitem{dai2018scancomplete}
A.~Dai, D.~Ritchie, M.~Bokeloh, S.~Reed, J.~Sturm, and M.~Nie{\ss}ner.
\newblock Scancomplete: Large-scale scene completion and semantic segmentation
  for 3d scans.
\newblock {\em arXiv preprint arXiv:1712.10215}, 2018.

\bibitem{drost20123d}
B.~Drost and S.~Ilic.
\newblock 3d object detection and localization using multimodal point pair
  features.
\newblock In {\em 2012 Second International Conference on 3D Imaging, Modeling,
  Processing, Visualization \& Transmission}, pages 9--16. IEEE, 2012.

\bibitem{gupta2015aligning}
S.~Gupta, P.~Arbel{\'a}ez, R.~Girshick, and J.~Malik.
\newblock Aligning {3D} models to {RGB-D} images of cluttered scenes.
\newblock In {\em Proceedings of the IEEE Conference on Computer Vision and
  Pattern Recognition}, pages 4731--4740, 2015.

\bibitem{hansen2009benchmarking}
N.~Hansen.
\newblock Benchmarking a bi-population cma-es on the bbob-2009 function
  testbed.
\newblock In {\em Proceedings of the 11th Annual Conference Companion on
  Genetic and Evolutionary Computation Conference: Late Breaking Papers}, pages
  2389--2396. ACM, 2009.

\bibitem{hansen2003reducing}
N.~Hansen, S.~D. M{\"u}ller, and P.~Koumoutsakos.
\newblock Reducing the time complexity of the derandomized evolution strategy
  with covariance matrix adaptation (cma-es).
\newblock {\em Evolutionary computation}, 11(1):1--18, 2003.

\bibitem{hua2016scenenn}
B.-S. Hua, Q.-H. Pham, D.~T. Nguyen, M.-K. Tran, L.-F. Yu, and S.-K. Yeung.
\newblock Scenenn: A scene meshes dataset with annotations.
\newblock In {\em 3D Vision (3DV), 2016 Fourth International Conference on},
  pages 92--101. IEEE, 2016.

\bibitem{huashrec2017}
B.-S. Hua, Q.-T. Truong, M.-K. Tran, Q.-H. Pham, A.~Kanezaki, T.~Lee,
  H.~Chiang, W.~Hsu, B.~Li, Y.~Lu, et~al.
\newblock Shrec’17: Rgb-d to cad retrieval with objectnn dataset.

\bibitem{huang2018holistic}
S.~Huang, S.~Qi, Y.~Zhu, Y.~Xiao, Y.~Xu, and S.-C. Zhu.
\newblock Holistic {3D} scene parsing and reconstruction from a single {RGB}
  image.
\newblock In {\em European Conference on Computer Vision}, pages 194--211.
  Springer, 2018.

\bibitem{izadi2011kinectfusion}
S.~Izadi, D.~Kim, O.~Hilliges, D.~Molyneaux, R.~Newcombe, P.~Kohli, J.~Shotton,
  S.~Hodges, D.~Freeman, A.~Davison, et~al.
\newblock Kinectfusion: real-time 3d reconstruction and interaction using a
  moving depth camera.
\newblock In {\em Proceedings of the 24th annual ACM symposium on User
  interface software and technology}, pages 559--568. ACM, 2011.

\bibitem{izadinia2017im2cad}
H.~Izadinia, Q.~Shan, and S.~M. Seitz.
\newblock Im2cad.
\newblock In {\em Computer Vision and Pattern Recognition (CVPR), 2017 IEEE
  Conference on}, pages 2422--2431. IEEE, 2017.

\bibitem{johnson1997spin}
A.~E. Johnson.
\newblock Spin-images: a representation for 3-d surface matching.
\newblock 1997.

\bibitem{kahler2015very}
O.~K{\"a}hler, V.~A. Prisacariu, C.~Y. Ren, X.~Sun, P.~Torr, and D.~Murray.
\newblock Very high frame rate volumetric integration of depth images on mobile
  devices.
\newblock {\em IEEE transactions on visualization and computer graphics},
  21(11):1241--1250, 2015.

\bibitem{keller2013real}
M.~Keller, D.~Lefloch, M.~Lambers, S.~Izadi, T.~Weyrich, and A.~Kolb.
\newblock Real-time 3d reconstruction in dynamic scenes using point-based
  fusion.
\newblock In {\em 3D Vision-3DV 2013, 2013 International Conference on}, pages
  1--8. IEEE, 2013.

\bibitem{kim2013guided}
Y.~M. Kim, N.~J. Mitra, Q.~Huang, and L.~Guibas.
\newblock Guided real-time scanning of indoor objects.
\newblock In {\em Computer Graphics Forum}, volume~32, pages 177--186. Wiley
  Online Library, 2013.

\bibitem{kim2012acquiring}
Y.~M. Kim, N.~J. Mitra, D.-M. Yan, and L.~Guibas.
\newblock Acquiring {3D} indoor environments with variability and repetition.
\newblock {\em ACM Transactions on Graphics (TOG)}, 31(6):138, 2012.

\bibitem{li2015database}
Y.~Li, A.~Dai, L.~Guibas, and M.~Nie{\ss}ner.
\newblock Database-assisted object retrieval for real-time {3D} reconstruction.
\newblock In {\em Computer Graphics Forum}, volume~34, pages 435--446. Wiley
  Online Library, 2015.

\bibitem{lim2013parsing}
J.~J. Lim, H.~Pirsiavash, and A.~Torralba.
\newblock Parsing ikea objects: Fine pose estimation.
\newblock In {\em Proceedings of the IEEE International Conference on Computer
  Vision}, pages 2992--2999, 2013.

\bibitem{mccormac2018fusion++}
J.~McCormac, R.~Clark, M.~Bloesch, A.~Davison, and S.~Leutenegger.
\newblock Fusion++: Volumetric object-level slam.
\newblock In {\em 2018 International Conference on 3D Vision (3DV)}, pages
  32--41. IEEE, 2018.

\bibitem{nan2012search}
L.~Nan, K.~Xie, and A.~Sharf.
\newblock A search-classify approach for cluttered indoor scene understanding.
\newblock {\em ACM Transactions on Graphics (TOG)}, 31(6):137, 2012.

\bibitem{newcombe2011kinectfusion}
R.~A. Newcombe, S.~Izadi, O.~Hilliges, D.~Molyneaux, D.~Kim, A.~J. Davison,
  P.~Kohi, J.~Shotton, S.~Hodges, and A.~Fitzgibbon.
\newblock Kinectfusion: Real-time dense surface mapping and tracking.
\newblock In {\em Mixed and augmented reality (ISMAR), 2011 10th IEEE
  international symposium on}, pages 127--136. IEEE, 2011.

\bibitem{niessner2013hashing}
M.~Nie{\ss}ner, M.~Zollh\"ofer, S.~Izadi, and M.~Stamminger.
\newblock Real-time 3d reconstruction at scale using voxel hashing.
\newblock {\em ACM Transactions on Graphics (TOG)}, 2013.

\bibitem{otsu1979threshold}
N.~Otsu.
\newblock A threshold selection method from gray-level histograms.
\newblock {\em IEEE transactions on systems, man, and cybernetics},
  9(1):62--66, 1979.

\bibitem{phamshrec2018}
Q.-H. Pham, M.-K. Tran, W.~Li, S.~Xiang, H.~Zhou, W.~Nie, A.~Liu, Y.~Su, M.-T.
  Tran, N.-M. Bui, et~al.
\newblock Shrec’18: Rgb-d object-to-cad retrieval.

\bibitem{rusu2009fast}
R.~B. Rusu, N.~Blodow, and M.~Beetz.
\newblock Fast point feature histograms (fpfh) for 3d registration.
\newblock In {\em Robotics and Automation, 2009. ICRA'09. IEEE International
  Conference on}, pages 3212--3217. Citeseer, 2009.

\bibitem{rusu20113d}
R.~B. Rusu and S.~Cousins.
\newblock 3d is here: Point cloud library (pcl).
\newblock In {\em Robotics and automation (ICRA), 2011 IEEE International
  Conference on}, pages 1--4. IEEE, 2011.

\bibitem{salas2013slam++}
R.~F. Salas-Moreno, R.~A. Newcombe, H.~Strasdat, P.~H. Kelly, and A.~J.
  Davison.
\newblock Slam++: Simultaneous localisation and mapping at the level of
  objects.
\newblock In {\em Proceedings of the IEEE conference on computer vision and
  pattern recognition}, pages 1352--1359, 2013.

\bibitem{salti2014shot}
S.~Salti, F.~Tombari, and L.~Di~Stefano.
\newblock Shot: Unique signatures of histograms for surface and texture
  description.
\newblock {\em Computer Vision and Image Understanding}, 125:251--264, 2014.

\bibitem{shao2012interactive}
T.~Shao, W.~Xu, K.~Zhou, J.~Wang, D.~Li, and B.~Guo.
\newblock An interactive approach to semantic modeling of indoor scenes with an
  {RGBD} camera.
\newblock {\em ACM Transactions on Graphics (TOG)}, 31(6):136, 2012.

\bibitem{song2017ssc}
S.~Song, F.~Yu, A.~Zeng, A.~X. Chang, M.~Savva, and T.~Funkhouser.
\newblock Semantic scene completion from a single depth image.
\newblock {\em Proceedings of 30th IEEE Conference on Computer Vision and
  Pattern Recognition}, 2017.

\bibitem{sun2018pix3d}
X.~Sun, J.~Wu, X.~Zhang, Z.~Zhang, C.~Zhang, T.~Xue, J.~B. Tenenbaum, and W.~T.
  Freeman.
\newblock {Pix3D}: Dataset and methods for single-image {3D} shape modeling.
\newblock In {\em Proceedings of the IEEE Conference on Computer Vision and
  Pattern Recognition}, pages 2974--2983, 2018.

\bibitem{tombari2010signature}
F.~Tombari, S.~Salti, and L.~Di~Stefano.
\newblock Unique signatures of histograms for local surface description.
\newblock In K.~Daniilidis, P.~Maragos, and N.~Paragios, editors, {\em Computer
  Vision -- ECCV 2010}, pages 356--369, Berlin, Heidelberg, 2010. Springer
  Berlin Heidelberg.

\bibitem{whelan2015elasticfusion}
T.~Whelan, S.~Leutenegger, R.~F. Salas-Moreno, B.~Glocker, and A.~J. Davison.
\newblock Elasticfusion: Dense slam without a pose graph.
\newblock {\em Proc. Robotics: Science and Systems, Rome, Italy}, 2015.

\bibitem{xiang2016objectnet3d}
Y.~Xiang, W.~Kim, W.~Chen, J.~Ji, C.~Choy, H.~Su, R.~Mottaghi, L.~Guibas, and
  S.~Savarese.
\newblock Objectnet3d: A large scale database for {3D} object recognition.
\newblock In {\em European Conference on Computer Vision}, pages 160--176.
  Springer, 2016.

\bibitem{xiang2014beyond}
Y.~Xiang, R.~Mottaghi, and S.~Savarese.
\newblock Beyond pascal: A benchmark for 3d object detection in the wild.
\newblock In {\em Applications of Computer Vision (WACV), 2014 IEEE Winter
  Conference on}, pages 75--82. IEEE, 2014.

\bibitem{zeng20173dmatch}
A.~Zeng, S.~Song, M.~Nie{\ss}ner, M.~Fisher, J.~Xiao, and T.~Funkhouser.
\newblock 3dmatch: Learning local geometric descriptors from rgb-d
  reconstructions.
\newblock In {\em Computer Vision and Pattern Recognition (CVPR), 2017 IEEE
  Conference on}, pages 199--208. IEEE, 2017.

\bibitem{zhou2018unsupervised}
X.~Zhou, A.~Karpur, C.~Gan, L.~Luo, and Q.~Huang.
\newblock Unsupervised domain adaptation for 3d keypoint estimation via view
  consistency.
\newblock In {\em Proceedings of the European Conference on Computer Vision
  (ECCV)}, pages 137--153, 2018.

\bibitem{zou2018complete}
C.~Zou, R.~Guo, Z.~Li, and D.~Hoiem.
\newblock Complete {3D} scene parsing from an {RGBD} image.
\newblock {\em International Journal of Computer Vision (IJCV)}, 2018.

\end{thebibliography}
}

\newpage
\begin{appendix}

%
In this appendix, we detail statistics regarding the Scan2CAD dataset in Sec.~\ref{sec:supp_dataset}.
In Sec.~\ref{sec:supp_metric}, we detail our evaluation metric for the alignment models.
We show additional details for our keypoint correspondence prediction network in Sec.~\ref{sec:supp_network} and we show example correspondence predictions.
We provide additional detail for our alignment algorithm in Sec~\ref{sec:supp_alignment}.
In Sec.~\ref{sec:supp_baselines}, we describe the implementation details of the baseline approaches.

\section{Dataset}
\label{sec:supp_dataset}
A compilation of our dataset is presented in \autoref{fig:dataset}. 
As a full coverage was aimed during the annotation, we can see the variety and richness of the aligned objects.

\begin{figure}[h]
\centering
\includegraphics[width=\linewidth]{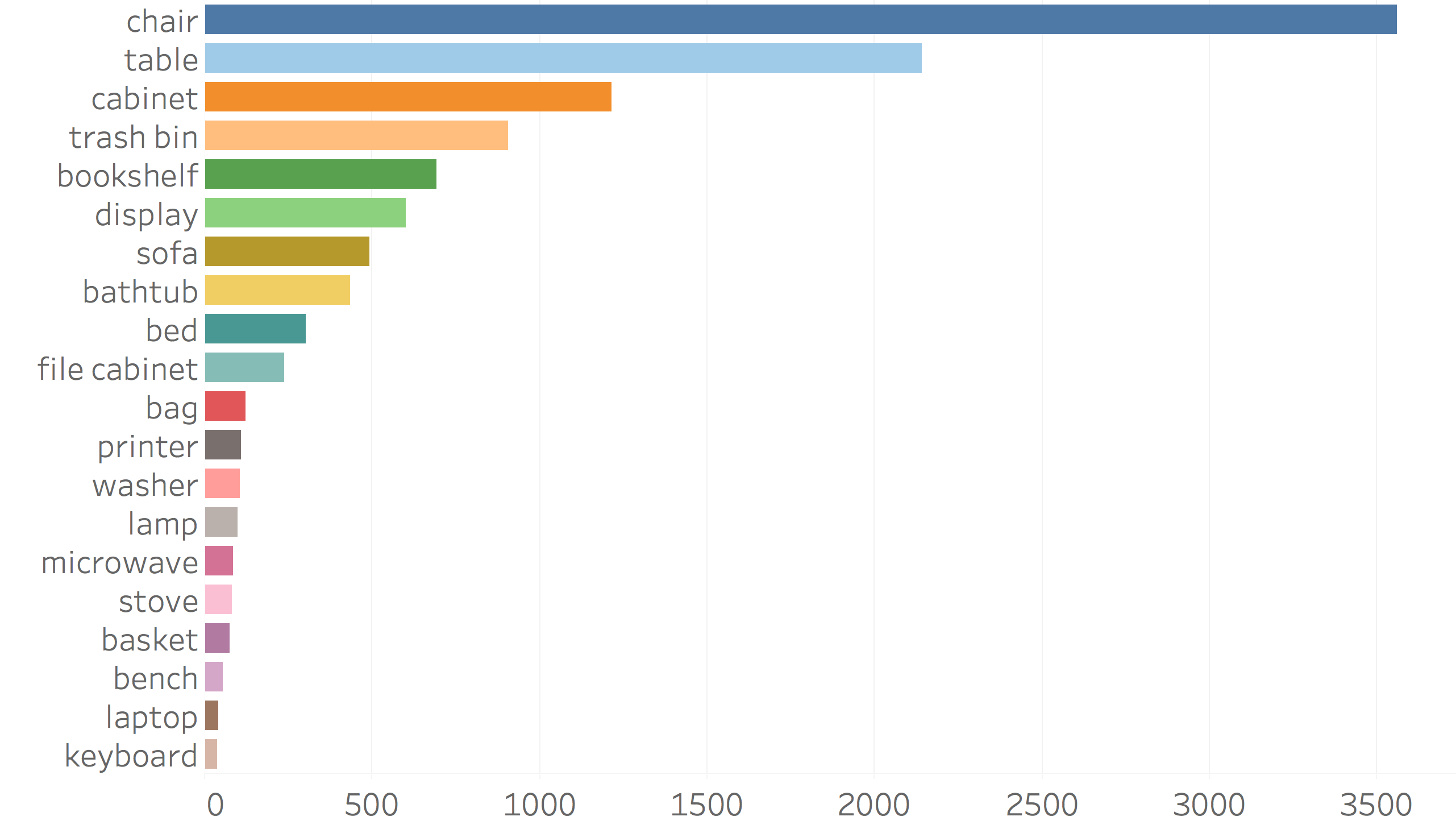}
\caption{Distribution of top 20 categories of annotated objects in our Scan2CAD dataset.}
\label{fig:category_distribution}
\end{figure}

\begin{figure}
\centering
\includegraphics[width=\linewidth]{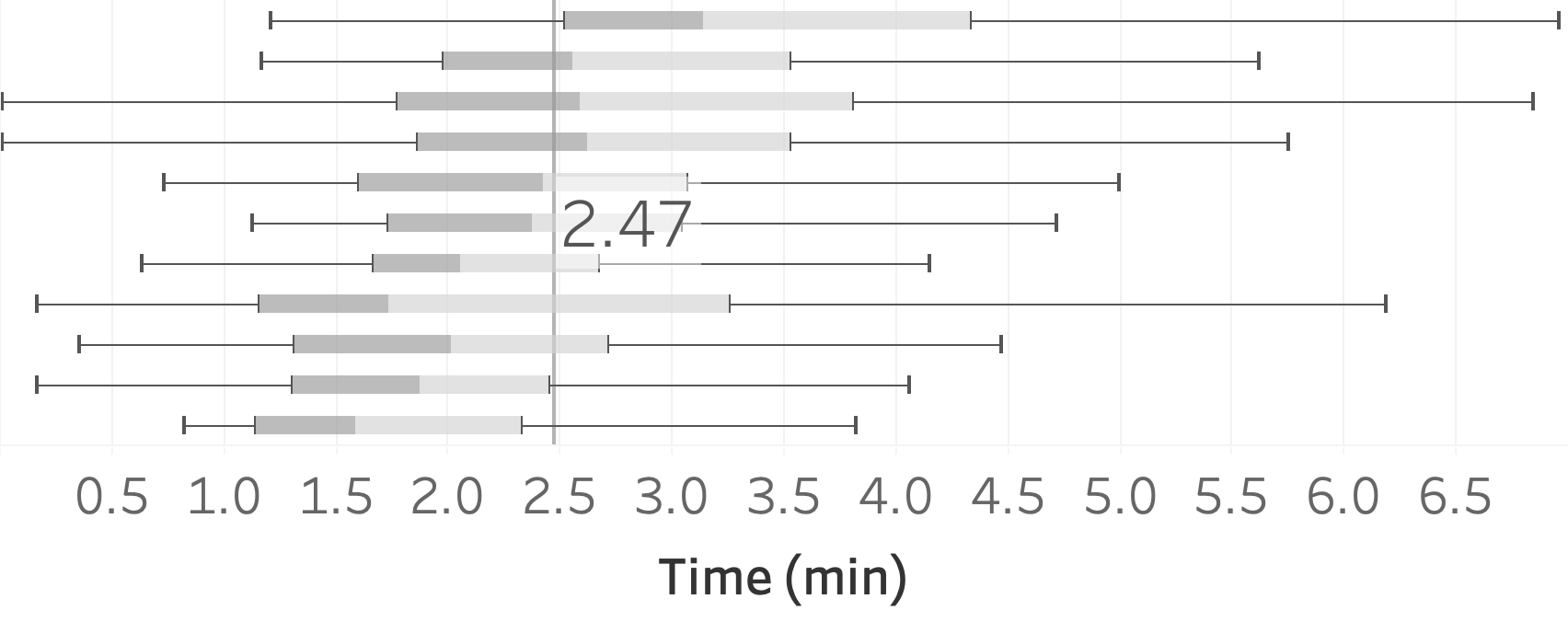}
\includegraphics[width=\linewidth]{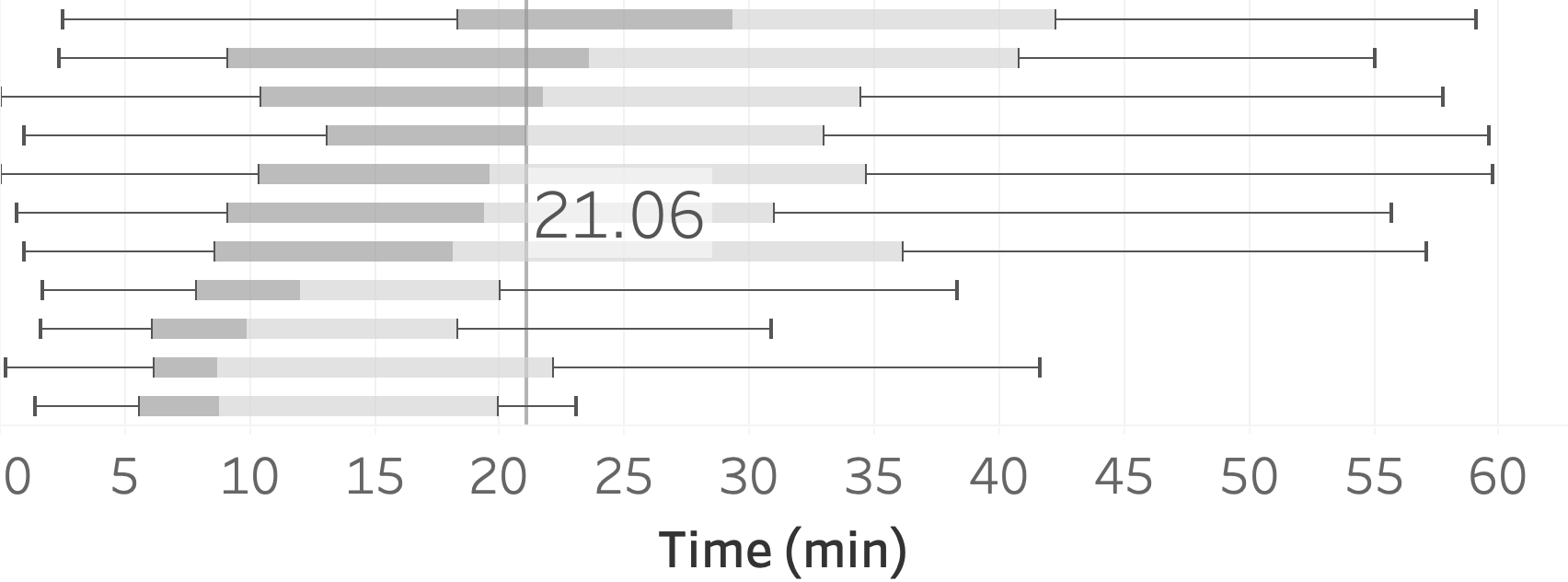}
\caption{Annotation timing distributions for each annotated object (top) and for each annotated scene (bottom). Each row shows a box-whisker plot with the median time and interquartile range for an annotator. The vertical rule shows the overall median across annotators.}
\label{fig:timings}
\end{figure}

\paragraph{Statistics}
 We show the object category statistics of our dataset in \autoref{fig:category_distribution}. Since our dataset is constructed on scans of indoor environments, it contains many furniture categories (e.g., chairs, tables, and sofas).
In addition, it also provides alignments for a wide range of other objects such as backpacks, keyboards, and monitors.

\paragraph{Timings} The annotation timings per object and per scan are illustrated in \autoref{fig:timings} (top) and \autoref{fig:timings} (bottom). 
On an object level, the timings are relatively consistent with little variance in time.
On a scan level, however, the variation in annotation time is larger which is due to variation in scene size.
Larger scenes are likely to contain more objects and hence require longer annotation times.

\begin{figure}[h!]
\centering
\includegraphics[width=\linewidth]{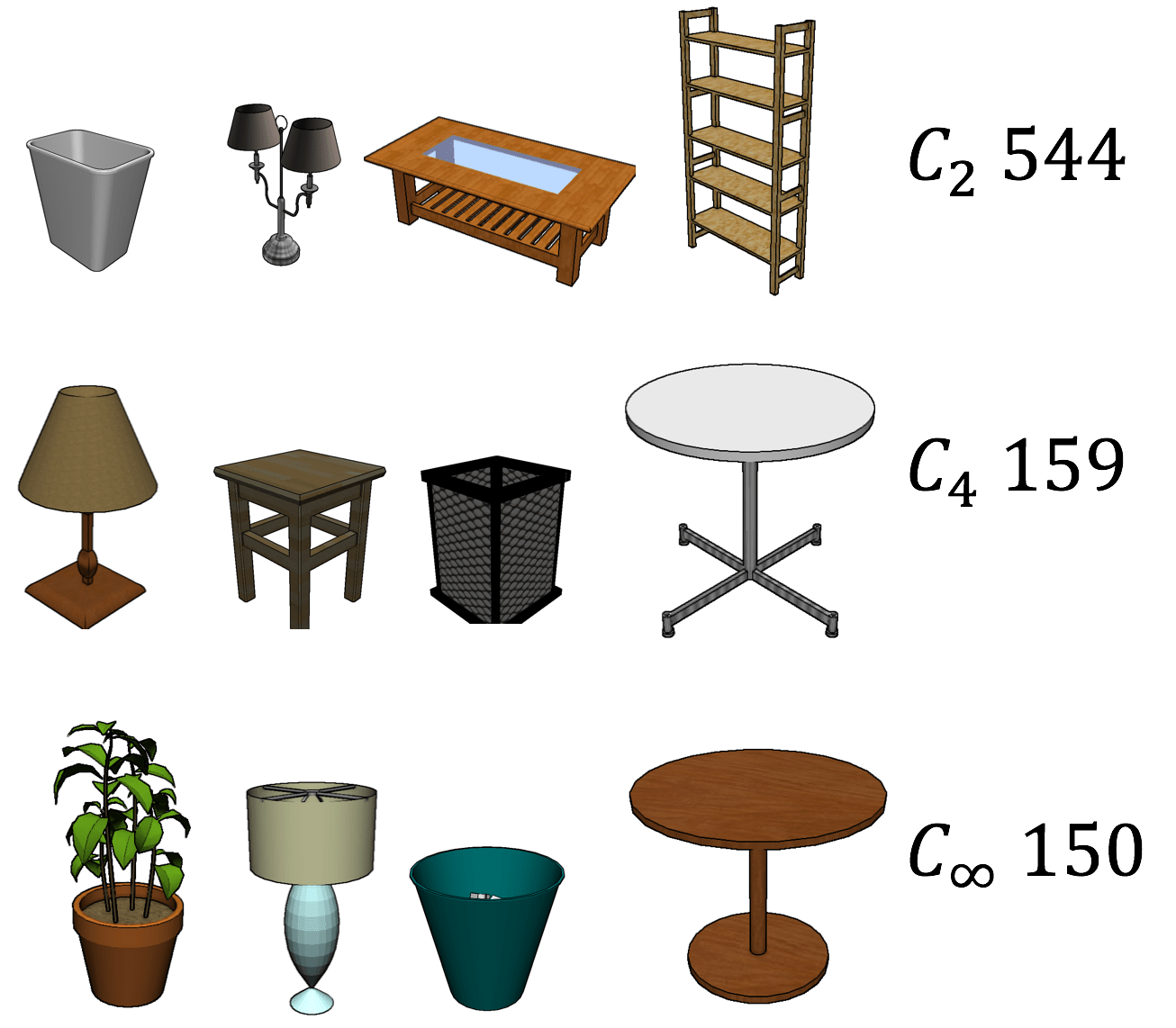}
\caption{Examples of symmetry annotations.}
\label{fig:symmetries}
\end{figure}

\paragraph{Symmetries}
In order to take into account the natural symmetries of many object categories during our training and evaluation, we collected a set of symmetry type annotations for all instances of CAD models.  \autoref{fig:symmetries} shows examples and total counts for all rotational symmetry annotations.

\section{Evaluation Metric}
\label{sec:supp_metric}
In this section, we describe the details of the algorithm for computing the alignment accuracy.  To compute the accuracy, we do a greedy matching of aligned CAD models to the ground truth CAD models.

\begin{algorithm}
\SetKwFunction{Union}{Union}\SetKwFunction{FunDist}{Distance}
 \KwData{1 \textbf{id-scan}, $N$ CADs (\textbf{id}, \textbf{cat}, \textbf{pose})}
 \KwResult{accuracy in $\%$}
 \KwSty{Init:} \\
 Get $N$ GT-CADs from database with \textit{key}=\textbf{id-scan}\\
 Set thresholds $t_t=20cm, t_r=20^{\circ}$, $t_s=20\%$\\
 counter = 0\;
 \For{c \textup{\textbf{in}} CADs}{
 \textbf{id}, \textbf{cat}, \textbf{pose} = c\\
  \For{c-gt \textup{\textbf{in}} GT-CADs}{
    \textbf{id}\textsubscript{GT}, \textbf{cat}\textsubscript{GT}, \textbf{pose}\textsubscript{GT} = c-gt\\
    \If{ \textbf{cat} == \textbf{cat}\textsubscript{GT}}{
        $\epsilon_t$ = \FunDist(\textbf{pose}.t, \textbf{pose}\textsubscript{GT}.t)\\
         $\epsilon_r$ = \FunDist(\textbf{pose}.r, \textbf{pose}\textsubscript{GT}.r, \textbf{sym}\textsubscript{GT})\\
          $\epsilon_s$ = \FunDist(\textbf{pose}.s, \textbf{pose}\textsubscript{GT}.s)\\
       \If{ $\epsilon_t$ $\leq$ $t_t$ and $\epsilon_r$ $\leq$ $t_r$ and $\epsilon_s$ $\leq$ $t_s$ }{
        counter ++\\
        remove \textbf{id}\textsubscript{GT} from GT-CADs\\
        break\\
    }
    }
    }
 }
 \KwOut{\textbf{accuracy} = counter/N}
 \vspace{0.5cm}
 \caption{Pseudo code of our evaluation benchmark. \textbf{id, cat, pose} denotes the id, category label and 9DoF alignment transformation for a particular CAD model. Note that the rotation distance function takes symmetries into account.}
 \label{alg:evaluation}
\end{algorithm}

For a given aligned scene \textbf{id-scan} with $N$ aligned CAD models, we query the ground truth alignment for the given scene. 
The evaluation script then iterates through all aligned candidate models and checks whether there is a ground truth CAD model of the same class where the alignment error is below the given bounds; if one is found, then the counter (of positive alignments) is incremented and the respective ground truth CAD model is removed from the ground truth pool.   See \autoref{alg:evaluation} for the pseudo-code.

\section{Correspondence Prediction Network}
\label{sec:supp_network}

\begin{figure}[htb!]
\begin{center}
   \includegraphics[width=0.8\linewidth]{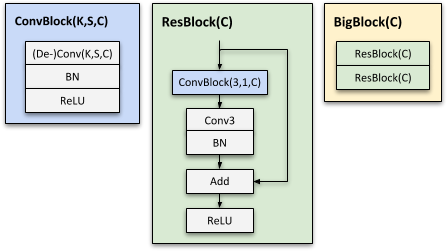}
    \end{center}
   \caption{CNN building blocks for our \OURS{} architecture. \textbf{K, S, C} stand for \textit{kernel-size}, \textit{stride} and \textit{num-channels} respectively.}
\label{fig:net_blocks}
\end{figure}

\paragraph{Network details} The details of the building blocks for our correspondence prediction network are depicted in \autoref{fig:net_blocks}.  See Figure 4 of the main paper for the full architecture.  We introduce the following blocks:
\begin{itemize}
    \item \textbf{ConvBlocks} are the most atomic blocks and consist of a sequence of \textbf{Conv3-BatchNorm-ReLU} layers as commonly found in other literature.
     \item \textbf{ResBlocks} are essentially residual skip connecting layers.
     \item \textbf{BigBlocks} contain two \textbf{ResBlocks} in succession.
\end{itemize}

\begin{figure}[htb!]
\centering
\includegraphics[width=\linewidth]{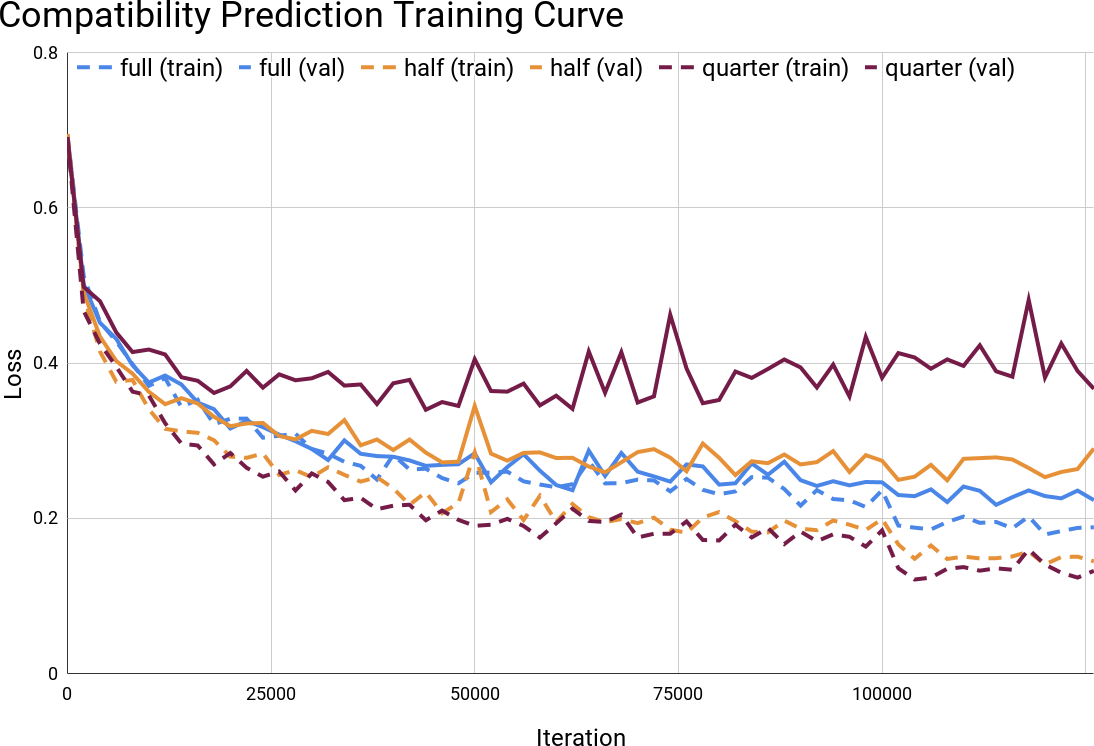}
\caption{Training and validation curves for varying training data sizes showing the probability score predictions. Experiments are carried out with full, half, and a quarter of the data set size. We see severe overfitting for half and quarter dataset training experiments, while our full training corpus mitigates overfitting.}
\label{fig:overfitting}
\end{figure}

\paragraph{Training curves}
\autoref{fig:overfitting} shows how much data is required for training the alignment approach.
The curves show predicted compatibility scores of our network.
We train our 3D CNN approach with different numbers of training samples (full, half and quarter of the dataset), and show both training and validation curves for each of the three experiments.
When using only a quarter or half of the dataset, we see severe overfitting.
This implies that our entire dataset provides significantly better generalization.

\begin{figure}[htb!]
\centering
\includegraphics[width=\linewidth]{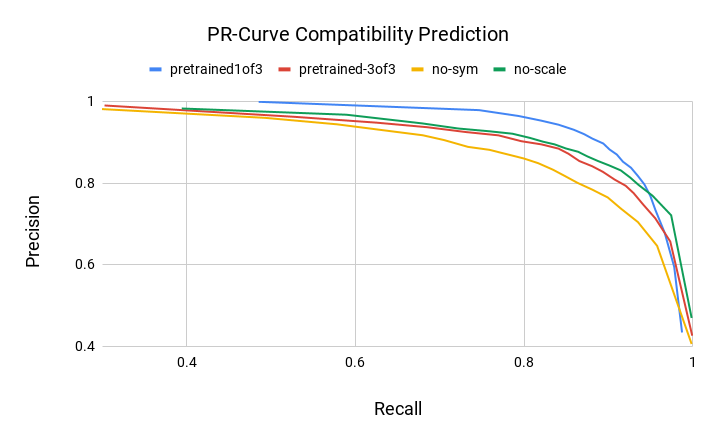}
\caption{Precision-recall curve of our compatibility score predictions.  }
\label{fig:prcurve_match}
\end{figure}

In \autoref{fig:prcurve_match}, we show the Precision-recall curve of the compatibility prediction of a our ablations (see Sec. 7.1 in the main paper). The PR-curves underline the strength of our best preforming network variation.

\paragraph{Correspondence predictions} Visual results of the correspondence prediction are shown in \autoref{fig:heatmaps}. 
One can see that our correspondence prediction network predicts as well symmetry-equivalent correspondences. 
The scan input with a voxel resolution of $3\text{cm}$ and a grid dimension of $64$ can cover $1.92\text{m}$ per dimension.
A larger receptive field is needed for large objects in order infer correspondences from a more global semantic context (see left-hand side first and second row.).

\section{Alignment Algorithm Details}
\label{sec:supp_alignment}
In order to remove misaligned objects, we prune objects after the alignment optimization based on the known free space of the given input scan.
This is particularly important for the unconstrained (`in-the-wild') scenario where the set of ground truth CAD models to be aligned is not given as part of the input.
For a given candidate transformation $T_m$ (as described in Sec. 6 in the main paper), we compute:

\begin{align*}
    &c = \frac{\sum_x^{\Omega^\text{occupied}_\text{CAD}} \mathcal{O}^\text{seen}_\text{scan} (T_{\text{world} \rightarrow \text{vox}, {\text{scan}}} \cdot T_m^{-1} \cdot T_{\text{vox} \rightarrow \text{world}, {\text{CAD}}} \cdot x )^2}{ | \Omega^\text{occupied}_\text{CAD} |} \notag \\
    &\Omega^{\text{occupied}}_{\text{CAD}} = \{x \in \Omega_{\text{CAD}} \mid  \mathcal{O}_\text{CAD}(x) < 1 \} \notag \\
    &\Omega^{\text{seen}}_{\text{scan}} = \{x \in \Omega_{\text{scan}} \mid  \mathcal{O}_\text{scan}(x) > - \tau \} \notag \\
    &\mathcal{O}^\text{seen}_\text{scan}(x) = \mathcal{O}_\text{scan}(x) \text{ if } x \in \Omega^{\text{seen}}_{\text{scan}} \text{ else } 0 \notag
\end{align*}
where $T_m^{-1}$ defines the transformation from CAD to scan, $\Omega$ defines a voxel grid space ($\subset \mathbb{N}^3$), $\tau$ is the truncation distance used in volumetric fusion (we use $\tau=15$cm), and $\mathcal{O}$ are look-ups into the signed distance function or distance functions for the scan or CAD model.
We also require that at least $30\%$ of the CAD surface voxels $\Omega^\text{occupied}_\text{CAD}$ project into seen space of the scan voxel grid $\Omega^{\text{seen}}_{\text{scan}}$.
Finally, we rank all alignments (of various models) per scene w.r.t. their confidence and prune all lower ranked models that are closer than $0.3$m to a higher ranked model.

\section{Alignment Error Analysis}
\begin{figure*}
\begin{center}
    \begin{subfigure}{\linewidth}
         \includegraphics[width=\linewidth]{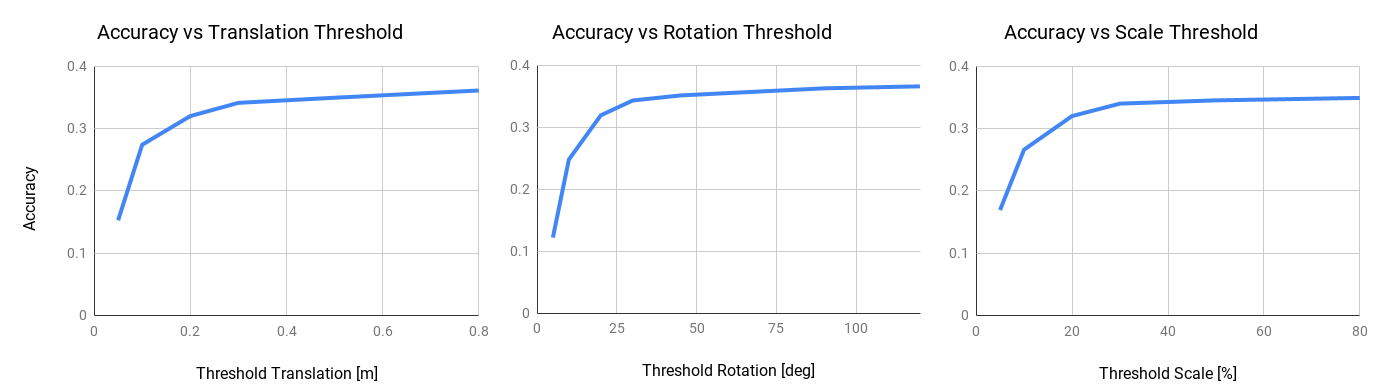}
    \end{subfigure}    
\end{center}
\vspace{-0.5cm}
\caption{Accuracy vs. varying thresholds for translation (left), rotation (middle) and scale (right). Only one threshold is varied whereas the remaining ones were held constant at their default value either $\epsilon_t = 0.2\text{m}, \epsilon_r = 20^\circ, \epsilon_s = 20\text{\%}$.}
\label{fig:accuracy_vs_thresholds}
\end{figure*}

\begin{figure*}[tbh!]
\centering
\includegraphics[width=\linewidth]{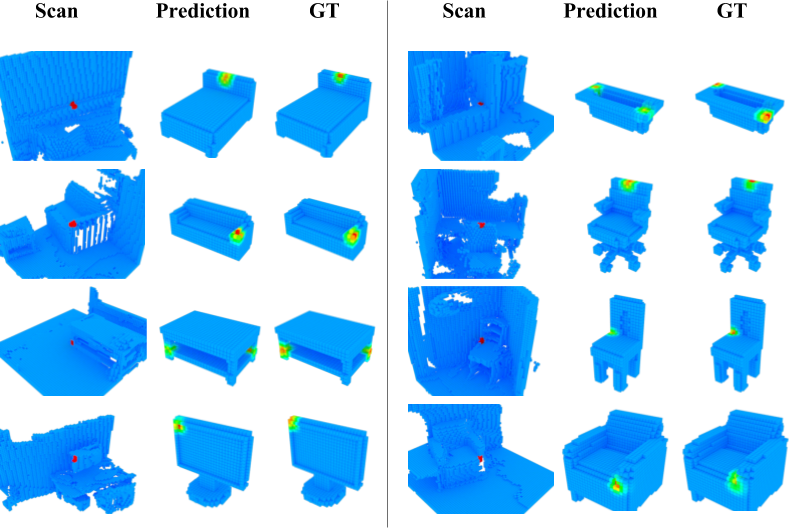}
\caption{Sample correspondence predictions over a range of various CAD models. Heatmaps contain symmetry-equivalent correspondences.}
\label{fig:heatmaps}
\end{figure*}

\begin{figure*}[tbh!]
\begin{center}
\end{center}
\includegraphics[width=\linewidth]{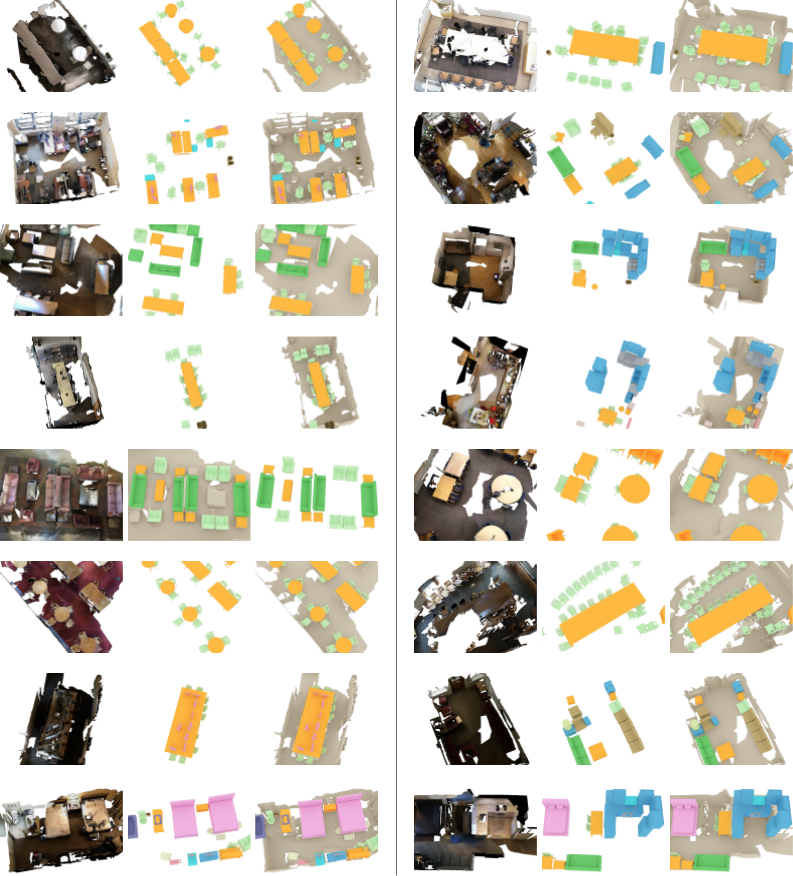}
\caption{Samples of annotated scenes. Left: 3D scan. Center: annotated CAD model arrangement; right: overlay CAD models onto scan.}
\label{fig:dataset}
\vspace{1cm}
\end{figure*}

Our alignment results have different sensibility for each parameter block (translation, rotation, scale). In order to gauge the stringency of each parameter block we varied the threshold for one parameter block and held the other two constant at the default value (see \autoref{fig:accuracy_vs_thresholds}). We observe that for the default thresholds $\epsilon_t = 0.2\text{m}, \epsilon_r = 20^\circ, \epsilon_s = 20\text{\%}$ all thresholds

\section{Baseline Method Details}
\label{sec:supp_baselines}
In the following, we provide additional details for the used baseline approaches. 
FPFH and SHOT work on point clouds and compute geometric properties between points within a support region around a keypoint. We use the implementation provided in the Point Cloud Library~\cite{rusu20113d}.

The method presented by Li~\etal~\cite{li2015database} takes the free space around a keypoint into account to compute a descriptor distance between a keypoint in scan and another keypoint in a CAD object. 
Here, we use the original implementation from the authors and modified it such that it works within a consistent evaluation framework together with the other methods. 
However, since we are not restricted to real-time constraints, we neglect the computation of the geometric primitives around the keypoints, which helps to find good initial rotation estimations. 
Instead, we computed all 36 rotation variants to find the smallest distance. 
We also replace the original 1-point RANSAC with another RANSAC as described below.

3DMatch~\cite{zeng20173dmatch} takes as input a 3D volumetric patch from a TDF around a keypoint and computes via a series of 3D convolutions and max-poolings a 512 dimensional feature vector. 
In order to train 3DMatch, we assemble a correspondence dataset as described in Sec. 5.3 in the main paper.
We train the network for 25 epochs using the original contrastive loss with a margin of 1. During test time, we extract the 3D patch around a detected Harris keypoint of both CAD object and scan and separately compute their feature vector.

For each method, we compute the feature descriptors for all keypoints in the scan and the CAD objects, respectively. 
We then find correspondences between pairs of keypoints if their height difference is less than $0.8m$ and if the L2 distance between the descriptors is below a certain threshold. 
Due to potential re-occurring structures in scan and CAD we select the top-8 correspondences with the smallest descriptor distances for each keypoint in the scan.

After establishing potential correspondences between the scan and a CAD object, we use a RANSAC outlier rejection method to filter out wrong correspondences and find a suitable transformation to align the CAD object within the scene. 
During each RANSAC iteration, we estimate the translation parameters and the up-right rotation by selecting 3 random correspondences. 
If the transformation estimate gives a higher number of inliers than previous estimates, we keep this transformation. 
The threshold of the Euclidean distance for which a correspondence is considered as an inlier is set to $0.20m$.
We use a fixed scale determined by the class average scale from our \OURS{} train set. 
For a given registration for a specific CAD model, we mark off all keypoints in the scan which were considered as inliers as well as all scan keypoints which are located inside the bounding box of the aligned CAD model. 
These marked keypoints will be ignored for the registration of later CAD models. 

To find optimal parameter for FPFH, SHOT, and Li~\etal, we construct an additional correspondence benchmark and ran a hyperparameter search based on the validation set.

\end{appendix}

\end{document}